\documentclass{article}

\usepackage[nonatbib, final]{neurips_2021}

\usepackage[utf8]{inputenc} 
\usepackage[T1]{fontenc}    
\usepackage{hyperref}       
\usepackage{url}            
\usepackage{booktabs}       
\usepackage{amsfonts}       
\usepackage{nicefrac}       
\usepackage{microtype}      
\usepackage{xcolor}         

\usepackage{amsmath, amsfonts, amsthm, amssymb}
\usepackage{bbm}
\usepackage{enumitem}
\usepackage{amsthm}
\newtheorem{assumption}{Assumption}
\newcommand{\indep}{\perp \!\!\! \perp}

\usepackage{subcaption}
\usepackage{graphicx}
\usepackage{wrapfig}
\usepackage{color, colortbl}
\usepackage[ruled, vlined]{algorithm2e}

\title{On Inductive Biases for Heterogeneous Treatment Effect Estimation}

\author{%
  Alicia Curth \\
  University of Cambridge\\
  \texttt{amc253@cam.ac.uk} \\
  \And
  Mihaela van der Schaar \\
  University of Cambridge \\
  University of California, Los Angeles \\
  The Alan Turing Institute\\
  \texttt{mv472@cam.ac.uk}
}

\begin{document}

\maketitle

\begin{abstract}
We investigate how to exploit structural similarities of an individual's potential outcomes (POs) under different treatments to obtain better estimates of conditional average treatment effects in finite samples. Especially when it is unknown whether a treatment has an effect at all, it is natural to hypothesize that the POs are similar -- yet, some existing strategies for treatment effect estimation employ regularization schemes that implicitly encourage heterogeneity even when it does not exist and fail to fully make use of shared structure. In this paper, we investigate and compare three end-to-end learning strategies to overcome this problem -- based on regularization, reparametrization and a flexible multi-task architecture -- each encoding \textit{inductive bias} favoring shared behavior across POs. To build understanding of their relative strengths, we implement all strategies using neural networks and conduct a wide range of semi-synthetic experiments. We observe that all three approaches can lead to substantial improvements upon numerous baselines and gain insight into performance differences across various experimental settings. 
\end{abstract}

\section{Introduction}
The advent of fields such as personalized medicine has led to rapid growth of the machine learning (ML) literature on heterogeneous treatment effect estimation in recent years \cite{hill2011bayesian, wager2018estimation, johansson2016learning, shalit2017estimating, yoon2018ganite, alaa2018limits, kunzel2019metalearners, nie2017quasi}. To further advance the understanding of how to incorporate insights from other areas of ML into treatment effect estimation, we revisit the well-established problem of estimating the conditional average treatment effect (CATE) of a binary treatment within the potential outcomes (PO) framework \cite{rubin2005causal}. The PO framework allows to conceptualize the problem as estimating the expected difference between an individual's expected `potential' outcome with and without treatment, of which only one is observed in the \textit{factual} world. This \textit{fundamental problem of causal inference} \cite{holland1986statistics} leads to the consensus that CATE estimation is not `just another' supervised learning problem \cite{alaa2018limits}. 

Under the standard assumption of ignorability -- which precludes hidden confounding -- we consider two \textit{statistical} features central to estimating CATE: (i) the presence of confounding and (ii) CATE being a \textit{contrast} between two PO functions, possibly exhibiting \textit{simpler} structure than each PO separately. Much of the recent ML literature on CATE estimation has focused on the first feature, and treated confounding as a \textit{covariate shift} problem. At this point, a range of sophisticated solutions exist which reduce the effect of confounding by balancing the covariate space \cite{johansson2016learning, shalit2017estimating}, importance weighting \cite{johansson2018learning, hassanpour2019counterfactual, hassanpour2020learning, assaad2020counterfactual} or propensity drop-out \cite{alaa2017deep}. How to exploit the second feature in an \textit{end-to-end manner}, however, has received little explicit attention so far and is what we focus on here.

We build on the intuition that the two tasks in the CATE problem -- estimating the expected PO with and without treatment -- are expected to be strongly related in practical applications (regardless of the\begin{wrapfigure}[10]{r}{0.45\textwidth}
    \centering
    \includegraphics[width=0.43\textwidth]{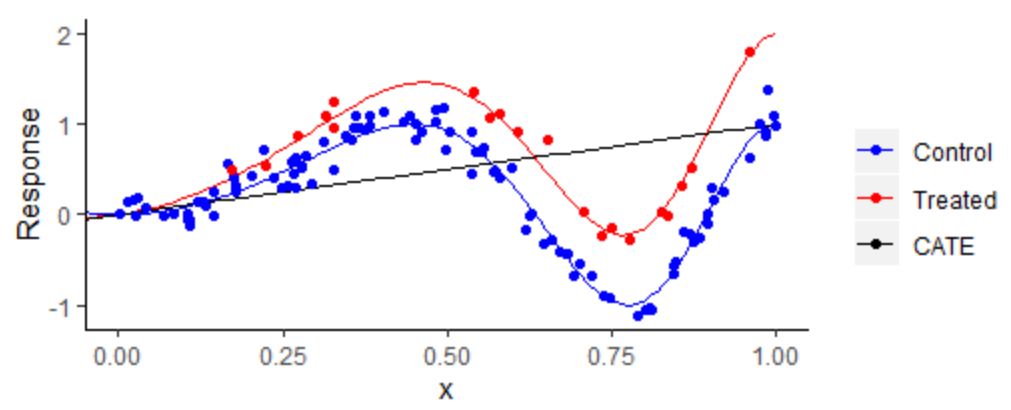}
      \vskip -0.05in
    \caption{Illustrative toy example: the treatment effect can be much simpler than each PO function separately}
 \label{fig:toyexample}
    \vskip -0.15in
\end{wrapfigure} presence of confounding). In fact, under the common null hypothesis of no treatment effect, we expect them to be identical. Even when there is a non-zero treatment effect, we expect shared structure: 
In medicine, for example, one distinguishes between biomarkers that are \textit{prognostic} of outcome  regardless of treatment status -- hence determining what is shared between the POs -- and biomarkers that are \textit{predictive} of treatment effectiveness -- determining heterogeneity \cite{ballman2015biomarker, sechidis2018distinguishing}. The induced difference between the POs is often expected to be small relative to the complexity of the PO functions themselves \cite{kunzel2019metalearners}, which, as in Fig. \ref{fig:toyexample}, could manifest in terms of CATE being a much simpler function than each PO.

One possible way of exploiting this is to estimate CATE directly using recently proposed  model-agnostic \textit{multi-stage} estimation strategies \cite{ kunzel2019metalearners, nie2017quasi, kennedy2020optimal, curth2020}. However, these estimators output an estimate of CATE only and do not include an estimate of the untreated PO, which is often of independent interest in practical decision support problems -- e.g. to trade off outcomes at baseline with the effectiveness of high-risk treatments \cite{coston2020counterfactual}. From this literature, we borrow the insight that explicitly \textit{targeting} learning strategies towards CATE, instead of only the POs, can lead to better estimators. However, instead of relying on multi-stage strategies, we operate within an \textit{end-to-end} learning framework similar to \cite{shalit2017estimating} and recent extensions. Due to a focus on confounding, this line of work did not explicitly investigate how to exploit the similarity of the PO functions beyond them sharing a jointly learned feature space. Combining the two lines of work, we investigate how to use end-to-end approaches to output better estimates of both the POs \textit{and} CATE by incorporating the assumption that the POs share much structure (which can result in a possibly simple CATE), as \textit{inductive bias}.

We focus on a fundamental question that has received little explicit attention so far: \textbf{\emph{How can we best exploit the structural similarities of the POs for CATE estimation?}} This question is crucial even in randomized experiments, making its solutions \textit{orthogonal} to any of the sophisticated strategies developed to handle confounding. Therefore, we investigate approaches exploiting the shared structure of the POs which can be applied to \textit{modify} existing modeling strategies, and thereby aim to provide guidance for \textit{improving existing} CATE estimators along a new dimension. As such, the goal of this paper is not to promote the use of a specific approach, method or architecture. Rather, we aim to build systematic understanding and greater intuition of the (dis)advantages of different approaches. This is crucial in the context of CATE estimation, where model selection is notoriously difficult due to the absence of ground truth treatment effects in practice. We focus on gaining insight into the effect of different approaches relying on \textit{the same} underlying ML method and use neural networks (NNs) due to their flexibility and popularity in related work, yet (variants of) the approaches we consider are applicable  to many likelihood- or loss-based ML methods.

\textbf{Contributions} We investigate three approaches incorporating inductive biases for shared structure into the estimation of the POs:  (1) a \textit{soft approach}, which relies on regularization to encourage the PO functions to be similar and is hence easy to combine with existing methods, (2) a \textit{hard approach}, which hardcodes an assumption on similarity into the model specification by reparametrization of the PO functions and (3) a \textit{flexible approach}, in which we build on ideas from multi-task learning to design a new architecture for CATE estimation (FlexTENet), which adaptively \textit{learns} what to share between the PO functions.  We implement instantiations of all approaches using NNs and evaluate their performance across a wide range of semi-synthetic experiments, varying in the structural similarity of the PO functions. We empirically confirm that all approaches can improve upon baselines, including both end-to-end and multi-stage approaches, and present a number of insights into the relative strengths of each approach.  We find that strategies significantly changing the model architecture -- hard and flexible approaches -- usually lead to the largest improvements, with FlexTENet performing best on average; yet even the simple soft approach often leads to notable performance increases -- an insight that can easily be incorporated into any existing method with treatment-specific parameters.

\section{Problem Definition and Key Challenges}
Assume we observe a sample $\mathcal{D}=\{(Y_i, X_i, W_i)\}^n_{i=1}$, with $(Y_i, X_i, W_i) \stackrel{i.i.d.}{\sim} \mathbbm{P}$. Here, $Y_i \in \mathcal{Y}$ is a continuous or binary outcome of interest, $X_i \in \mathcal{X} \subset \mathbbm{R}^d$ a vector of possible confounders (i.e. pre-treatment covariates) and $W_i\in \{0, 1\}$ is a binary treatment, assigned according to propensity score $\pi(x) = \mathbbm{P}(W=1| X=x)$. Using the Neyman-Rubin potential outcomes (PO) framework \cite{rubin2005causal}, our main interest lies in the individualized treatment effect: the difference between the PO  $Y_i(1)$ if treatment is administered ($W_i=1$) and $Y_i(0)$ if individual $i$ is not treated ($W_i=0$).  However, only one of the POs is observed as $Y_i = W_i Y_i(1) + (1-W_i)Y_i(0)$. Therefore, we focus on estimating the conditional average treatment effect (CATE)
\begin{equation}
\textstyle \tau(x) = \mathbbm{E}[Y(1) - Y(0) |X =x]
\end{equation}
which is the expected treatment effect for an individual with covariate values $X=x$. We operate under the standard identifying assumptions in the PO framework: 

\begin{assumption}\label{causass}[Consistency, unconfoundedness and overlap]
Consistency: If individual i is assigned treatment $w_i$, we observe the associated potential outcome $Y_i=Y_i(w_i)$. 
Unconfoundedness: there are no unobserved confounders, so that $Y(0), Y(1) \indep W | X$. Overlap: treatment assignment is non-deterministic, i.e. $0 < \pi(x) < 1 \text{, }\forall x\in \mathcal{X}$. 
\end{assumption}

\subsection{The Key Challenges of CATE Estimation}
As the ability to interpret a treatment effect estimate as causal ultimately relies on a set of \textit{untestable} assumptions, the \textit{unique} difficulty in making causal claims lies in using domain expertise to argue whether a treatment effect is identifiable \cite{pearl2009causality}. Given identifiability, CATE estimation is a purely statistical problem -- thus, if one is willing to rely on assumption \ref{causass}, CATE can be estimated using observed data. A simple strategy for doing so (also known as the T-learner \cite{kunzel2019metalearners}) obtains regression estimates $\hat{\mu}_w(x)$ of $\mu_w(x)=\mathbb{E}[Y|X=x, W=w]$,  applying standard supervised learning methods using only observed data for which $W=w$, and finally sets $\hat{\tau}(x) = \hat{\mu}_1(x) - \hat{\mu}_0(x)$. Yet, this seemingly straightforward solution is oblivious to two \textit{statistical} challenges of CATE estimation: 

 \textbf{1. Confounding}: If $\pi(x)$ is not constant, then the distribution of covariates in treatment and control groups differs. Such imbalance can be the result of confounders, which are variables that affect both treatment selection and outcomes, and can be problematic when the PO functions are fit on the factual data using empirical risk minimization (ERM) because each problem is solved with respect to the wrong empirical distribution -- namely $X \sim \mathbbm{P}(\cdot|W=w)$ instead of $X \sim \mathbbm{P}(\cdot)$. 
 While this problem is not unique to CATE estimation -- it is equivalent to the covariate shift problem encountered in e.g. domain adaptation \cite{shimodaira2000improving} -- it is usually emphasized as one of its main difficulties and motivated the literature on balanced representation learning \cite{johansson2016learning, shalit2017estimating}.

\textbf{2. CATE is the difference between two functions: } While supervised learning usually targets a single function, the goal of CATE estimation is to estimate \textit{the difference} between two related functions most accurately -- which may require different considerations than estimating each function separately. 
To see this, consider the MSE of estimating CATE and let $\textstyle{\epsilon_{sq}(\hat{f}(x)) = \mathbbm{E}_{X\sim \mathbbm{P}}[(\hat{f}(X)\!-\!f(X))^2]}$ denote the MSE for an estimate $\textstyle \hat{f}(x)$ of $\textstyle f(x)$. If we simply were to estimate $\textstyle \hat{\tau}(x) = \hat{\mu}_1(x)\!-\!\hat{\mu}_0(x)$ as the difference between two \textit{separately} learned functions, we would have that (up to constants) $ \textstyle   \epsilon_{sq}(\hat{\tau}(x)) \lesssim  \epsilon_{sq}(\hat{\mu}_1(x))\! + \!\epsilon_{sq}(\hat{\mu}_0(x)) \lesssim Rate_{\mu_{1}} \!+\! Rate_{\mu_{0}}$.
The convergence rates $Rate_{\mu_w}$ depend on the used estimator and assumptions on e.g. smoothness or sparsity of the PO functions; a well-known example would be  \cite{stone1980optimal}'s nonparametric minimax rate. If we had oracle access to both POs and could regress $Y(1)\!-\!Y(0)$ on $X$ directly, we would have $\textstyle \epsilon_{sq}(\hat{\tau}(x)) \lesssim Rate_\tau$.  The assumption that $\tau(x)$ is often much simpler than each $\textstyle \mu_w(x)$ separately \cite{kunzel2019metalearners} translates into $\textstyle Rate_\tau < \max_w Rate_{\mu_w}$, highlighting that targeting CATE directly could lead to faster convergence. Similarly, any shared structure across the PO regression tasks could also be exploited to improve upon the simple additive bound above. These observations motivated much of this paper, as they have largely been neglected in the literature on end-to-end learning for CATE. They have, however, been the motivation for some model-agnostic multi-stage learners which we discuss next. 

\section{Related Work}
\textbf{Direct and indirect meta-learners for CATE} Recent literature has developed a number of \textit{model-agnostic}  learning strategies for CATE estimation (also known as `meta-learners' \cite{kunzel2019metalearners}). Within this class, we consider an \textit{indirect} estimator any strategy that uses observed data to obtain regression estimates of the PO functions $\hat{\mu}_w(x)$ and then sets $\hat{\tau}(x)=\hat{\mu}_1(x)-\hat{\mu}_0(x)$. This includes \cite{kunzel2019metalearners}'s model-agnostic S- and T-learner; additionally the majority of model-specific ML-based CATE estimators also follow an indirect strategy (including all NN-based estimators we discuss below). Conversely, we consider learners \textit{direct} estimators if they target $\tau(x)$ directly. As $Y(1)-Y(0)$ is unobserved, multiple existing strategies construct \textit{pseudo-outcomes} $Y_\eta$ for which it holds that $\mathbbm{E}[Y_\eta|X=x]=\tau(x)$ for some nuisance parameters $\eta(x)$ which can be estimated from observational data. Different learners require estimation of different nuisance parameters, which often include propensity score $\pi(x)$ and/or PO functions $\mu_w(x)$. All direct estimation strategies that we are aware of -- X-learner \cite{kunzel2019metalearners}, R-learner \cite{nie2017quasi}, DR-learner \cite{kennedy2020optimal}, PW-learner and RA-learner \cite{curth2020} -- proceed in a two-stage manner: they first obtain plug-in nuisance parameter estimates $\hat{\eta}(x)$, and then estimate $\tau(x)$ by regressing $Y_{\hat{\eta}}$ on $X$. Under some conditions, these two-stage learners can attain the oracle rate $Rate_\tau$. We give a more detailed overview of all meta-learner strategies in appendix A.

\textbf{NN-based CATE estimators}
Complementing model-agnostic strategies, many adaptations of specific ML methods for CATE estimation have been proposed recently. We build on NN-based estimation strategies due to their flexibility and popularity in related work\footnote{As we discuss in Appendix A, other popular estimators, which we do not consider further as we are interested in the effect of different approaches relying on \textit{the same} underlying ML method, are based on Generalized Random Forests \cite{athey2019generalized}, Bayesian Additive Regression Trees \cite{hill2011bayesian, hahn2017bayesian}, Gaussian Processes \cite{alaa2017bayesian, alaa2018limits} and GANs \cite{yoon2018ganite}.}. Much work on CATE estimation using NNs has focused on handling confounding, most prominently by learning shared and balanced feature representations for the two PO functions \cite{johansson2016learning, shalit2017estimating}.  Formally, this strategy entails jointly learning a shared feature map, and two PO-specific regression heads (fit using only the data of the corresponding treatment group) each parametrized by a NN. The output heads are then used for indirect estimation of CATE. Without further regularization, this leads to the TARNet specification, while CFRNet introduces a regularization term which encourages the network to learn representations that are balanced, i.e. have indistinguishable distributions across treatment groups \cite{shalit2017estimating}. Recent extensions investigated incorporating weighting strategies as an additional remedy for confounding into this framework \cite{johansson2018learning, hassanpour2019counterfactual, hassanpour2020learning, assaad2020counterfactual}, considered targeting towards \textit{average} treatment effects \cite{shi2019adapting, hatt2021estimating} or allowed for both shared \textit{and} private feature spaces for the PO functions \cite{curth2020}.  

\textbf{Relationship to multi-task learning} The architectures proposed in \cite{shalit2017estimating} and extensions effectively take a multi-task learning (MTL) approach to PO estimation, relying on \textit{hard parameter sharing} \cite{caruana1997multitask} in the first $d_r$ layers of the used network, and no sharing in the top $d_h$  layers.  While sharing a feature space will lead to some shared behavior between the estimated PO functions, it does not allow to fully exploit underlying similarity -- e.g. if there are purely prognostic effects it might be better to share \textit{some} information also between top layers. Below, we therefore investigate strategies to incorporate (additional) inductive bias for shared behavior into PO estimation. These are inspired by work in transfer learning \cite{xuhong2018explicit}, domain adaptation \cite{daume2009frustratingly, bousmalis2016domain} and, in particular, MTL \cite{evgeniou2005learning, misra2016cross, ruder2017learning}, all allowing for flexible modeling of shared and task-specific aspects of a problem. Note, however, that MTL and CATE estimation have distinct statistical target parameters and goals -- MTL is concerned with achieving a good \textit{average performance in prediction of outcomes} across tasks, while the main target of CATE estimation is \textit{estimating the expected difference between outcomes} -- making it non-obvious a priori whether methods successful in the former will perform well in the latter problem.

\section{Inductive Biases for CATE Estimation}\label{main}
In this section, we consider end-to-end approaches for incorporating the prior belief that $\mu_0(x)$ and $\mu_1(x)$ will share much structure (which can imply that $\tau(x)$ is simpler than $\mu_w(x)$)  as inductive bias. Here, we define inductive bias as the mechanism by which some hypothesis functions are preferred over others during learning \cite{haussler1988quantifying}. Throughout and for the remainder of this paper, we focus on exploiting the expected similarity between the PO functions, and disregard the impact of confounding on ERM -- with the understanding that existing strategies, such as balancing or weighting, are orthogonal solutions that could readily be applied to complement any strategy we discuss.

\subsection{Implicit inductive biases in indirect learners}
We begin by examining the nature of the inductive biases present in popular indirect learners: a T-learner based on vanilla NNs (TNet) and TARNet. Recall that TARNet jointly learns a representation $\Phi: \mathcal{X} \rightarrow \mathcal{S}$, parametrized by a dense NN with $d_r$ layers and $n_r$ hidden units,  and regression heads $h_w: \mathcal{S} \rightarrow \mathcal{Y}$, each parametrized by a dense NN with $d_h$ layers and $n_h$ hidden units. A TNet can be seen as a special case of TARNet with $\Phi(x)=x$, i.e. no joint learning of feature spaces\footnote{In our implementations, to give similar capacity to the resulting PO estimators and in analogy with a `no parameter sharing' strategy, we give each $h_w$ in a TNet additional access to $d_r$ layers with $n_r$ hidden units.}. Below, we will refer to the weights in $h_w$ and $\Phi$ as $\Theta_{h_w}$ and $\Theta_{\Phi}$, respectively. During training, both TARNet and TNet  use a loss function that takes the general form
\begin{equation}\label{standardloss}
  \textstyle \mathcal{L}_F + \lambda_1 \sum_{w \in \{0, 1\}} \mathcal{R}(\Theta_{h_w})\end{equation}
where $\mathcal{L}_F = \sum^n_{i=1} l(y_i, h_{w_i}(\Phi(x_i)))$, $\mathcal{R}$ is a regularizer (usually L2), which regularizes the complexity of each $h_w$ separately, and we dropped the regularization term for the shared representation -- $\lambda_1 \mathcal{R}(\Theta_\Phi)$ -- from the equation as it is not of relevance for the following discussions.

While the PO functions can thus share a jointly learned feature space, they are \textit{not} encouraged to be similar beyond this. Instead, they are \textit{separately} regularized to be a simple function, leading to their difference -- $\tau(x)$ -- being highly instable and hence to an implicit inductive bias that encourages treatment effect heterogeneity a priori (see also the discussions of this phenomenon in \cite{nie2017quasi, hahn2017bayesian}). This is neither in line with a scientific null hypothesis of no treatment effect (heterogeneity) nor the assumption that  $\mu_1(x)$ and $\mu_0(x)$ should be close due to the existence of prognostic effects. The instability of $\hat{\tau}(x)$ can also be seen as a consequence of indirect learners not being well \textit{targeted} towards CATE. In fact, using this regularization scheme, it is not even possible to control the complexity of CATE directly.  

\begin{figure}[t]
    \centering
    \includegraphics[width=\textwidth]{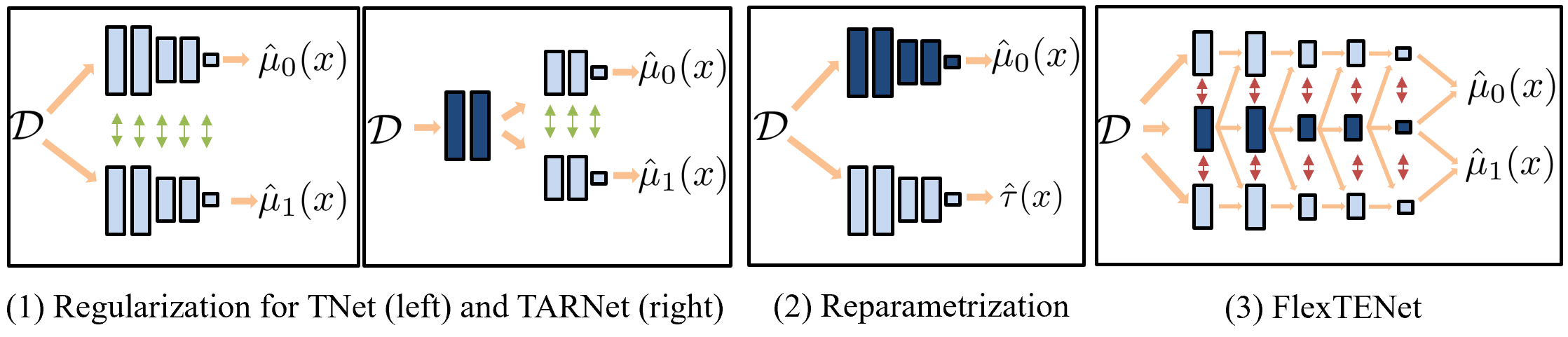}
    \vskip -0.05in
    \caption{The three approaches under investigation. Dark layers indicate parameters shared between POs, light layers indicate private parameters. Green arrows indicate regularization encouraging parameters to be similar, red arrows indicate regularization that encourages orthogonalization.}
    \label{fig:strategies}
\end{figure}
\subsection{Explicit inductive biases for CATE estimation}
We investigate three approaches modifying existing end-to-end learners to encourage shared structure in the POs and hence incorporate an inductive bias for simpler $\tau(x)$. Ordered by ease of implementation, we distinguish between (1) a soft approach relying on regularization, (2) a hard approach based on reparametrization and (3) a flexible approach (FlexTENet) which explicitly learns what to share between the POs. The architectures for each approach are depicted in Figure \ref{fig:strategies}. Note that although all approaches are targeted at estimating CATE, only (2) directly outputs an estimate of $\tau(x)$. 

\paragraph{Soft approach -- Regularization}
The most straightforward strategy to fixing the regularization-induced inductive bias towards heterogeneity discussed above would be to simply change how the PO functions are regularized. Instead of regularizing them separately, one could regularize the difference between the weights in the output heads -- which ultimately determines $\tau(x)$ --, corresponding to an inductive bias towards small treatment effect heterogeneity\footnote{Note that, by convention, we regularize only the weights of NNs, and \textit{not} biases (offsets), resulting in penalization only of non-constant treatment effects.}. Analogously to (\ref{standardloss}), this leads to a loss
\begin{equation}\label{regloss}
 \textstyle   \mathcal{L}_F + \lambda_1 \mathcal{R}(\Theta_{h_0}) + \lambda_2 \mathcal{R}(\Theta_{h_1}-\Theta_{h_0})
\end{equation}
Choosing $\lambda_2 > \lambda_1$ additionally reinforces the inductive bias towards simple $\tau(x)$. We further discuss how to set hyperparameters such as $\lambda_2$ (which is shared by all considered approaches) in Appendix B.1. This regularization-based approach is attractive because it is extremely easy to implement, is directly applicable to any loss-based method with treatment-group-specific parameters, has intuitive appeal and does not heavily constrain the functions the hypotheses are able to represent. At the same time, the latter point is a downside of this approach, since this might also result in only marginal gains. 

\paragraph{Hard approach -- Reparametrization} Instead of regularizing the difference between the PO functions, we could also \textit{reparametrize} our estimators, a strategy that e.g. \cite{imai2013estimating}'s LASSO for CATE estimation and \cite{hahn2017bayesian}'s Bayesian Causal Forest rely on. Instead of estimating $\mu_0(x)$ and $\mu_1(x)$ separately, we can build on the identity $\mu_1(x) = \mu_0(x) + \tau(x)$ and effectively estimate $\tau(x)$ directly as an offset from $h_0(x)$, parametrized by a NN $h_\tau(x)$ with weights $\Theta_{h_\tau}$. This leads to a loss  
\begin{equation}
\textstyle    \mathcal{L}_F + \lambda_1 \mathcal{R}(\Theta_{h_0}) + \lambda_2 \mathcal{R}(\Theta_{h_\tau})
\end{equation}
with $\mathcal{L}_F=\sum^n_{i=1} l(y_i, h_{0}(x_i) + w_i h_\tau(x_i))$ for continuous $y$, which seems analogous to (\ref{regloss}) but parametrizes $\tau(x)$ explicitly -- giving the investigator control over the complexity of $\tau(x)$ \textit{directly} -- and \textit{hard-codes} the assumption that the shared structure between the POs is \textit{additive}. This approach may be at a disadvantage if $\mu_1(x)$ is \textit{simpler} than $\mu_0(x)$, since one would then be better off using the reverse parametrization $\mu_0(x)=\mu_1(x)-\tau(x)$. More generally,  it is also possible that the relationship between $\mu_0(x)$ and $\mu_1(x)$ is not additive, e.g. if $\mu_1(x)$ is a simple transformation of $\mu_0(x)$ -- say a multiplicative or logarithmic transformation (as is the case in the popular IHDP benchmark \cite{hill2011bayesian}) -- such that a different parametrization would lead to an easier learning problem.

\paragraph{Flexible approach -- FlexTENet} Unfortunately, the parametrization leading to the easiest learning problem is usually not known in practice. Alternatively, one could thus rely on a strategy that can automatically and flexibly \textit{learn} which information to (hard-)share between the PO functions. Inspired by architectures in MTL and domain adaptation that \textit{explicitly} anticipate shared and private structure \cite{daume2009frustratingly, bousmalis2016domain, ruder2017learning}, we therefore propose a new architecture for treatment effect estimation, FlexTENet (Flexible Treatment Effect Network), as a final strategy. As depicted in Fig. \ref{fig:strategies}, it has private ($\mu_w(x)$-specific) subspaces -- which ultimately determine $\tau(x)$ -- and a shared subspace in each layer (including the output layer), allowing the model to automatically learn which information to share \textit{at each layer} of the network.
In principle, any MTL method could be adapted for PO estimation, yet, as we discuss in appendix B.2, we propose the FlexTENet specification because it \textit{generalizes} many existing strategies. Given its flexibility, we expect that such a general architecture should perform well \textit{on average} -- an appealing feature given that model-selection is nontrivial in CATE problems.

We implement FlexTENet using a specification matching TARNet to allow for direct comparisons, and consider $d_r+d_h$ layers, within which each private and shared subspace has $n_{k, p}$ and $n_{k, s}$ hidden units, respectively, where we let $n_{k,p}=n_{k, s}=\frac{1}{2}n_k$, $k \in \{r, h\}$, for simplicity.  For layer $l > 1$, let $m^{l-1}_p$ and $m^{l-1}_s$ denote the output dimensions of shared and private subspace of the previous layer, let  $\theta^l_s \in \mathbbm{R}^{m^{l-1}_s \times m^{l}_s}$ denote the weights in the shared subspace, while $\theta^l_{p_w} \in \mathbbm{R}^{(m^{l-1}_s + m^{l-1}_p) \times m^{l}_p}$ denotes the weights in each private subspace\footnote{The difference in input dimensions arises as we only allow communication from shared subspaces to private subspaces and not the reverse; refer to Appendix B.3 for pseudocode of a FlexTENet forward pass.}.  To discourage redundancy and encourage identification of private structure, we apply regularizers to orthogonalize the shared and private subspaces.  Like \cite{bousmalis2016domain, ruder2017learning} we rely on \cite{salzmann2010factorized}'s orthogonal regularizer $\textstyle \mathcal{R}_o(\Theta_s,\Theta_{p_0}, \Theta_{p_1}) = \sum_{w \in \{0, 1\}}\sum^L_{l=1} \lVert {\theta^l_s}^\top \theta^l_{p_w, 1:m^{l-1}_s} \rVert^2_{F}$ where $\lVert \cdot \rVert^2_{F}$ denotes the squared Frobenius norm. This leads to the following loss function
\begin{equation}
 \textstyle \mathcal{L}_F\! +\!\lambda_1\mathcal{R}(\Theta_{s})\!+\! \lambda_2\!\sum_{w \in \{0, 1\}}\mathcal{R}(\Theta_{p_w})\!+\!\lambda_o \mathcal{R}_o(\Theta_s,\Theta_{p_0}, \Theta_{p_1})\!
\end{equation}
where setting $\lambda_2 > \lambda_1$ adds inductive bias encouraging the shared space to be used first.

\subsubsection{Underlying assumptions and theory}
 \textbf{Which assumptions do these approaches encode?} Motivated by real-world applications in which prognostic effects are often assumed stronger than predictive ones \cite{kunzel2019metalearners, hahn2017bayesian}, our central assumption is \textit{`there is much shared structure between $\mu_0(x)$ and $\mu_1(x)$'}. This assumption is purposefully abstract, allowing it to manifest as different specific assumptions depending on the used ML method\footnote{While both hard and soft approach are directly applicable to any loss-based method, our flexible approach is specific to NNs. When using a different ML method, similar strategies could be constructed by adapting (i) existing MTL approaches or (ii) a model-agnostic approach similar to \cite{daume2009frustratingly}, creating PO-specific feature spaces.} and regularizer ($\mathcal{R}$); e.g. in regression with L0-penalty, $\tau(x)$ would be assumed linear and sparse, while in our case, using NNs with L2-penalty, the $\mu_w(x)$ are implicitly assumed to be close in some function class, with smooth differences. We thus consider inductive biases for $\tau(x)$ \textit{relative} to the inductive bias in the original method used to estimate the $\mu_w(x)$; we effectively investigate how to best re-target these biases to control $\tau(x)$ explicitly. Further, there is a conceptual difference between assumptions in soft \& flexible and hard approach; only the latter assumes shared structure to be additive. 
 
 \textbf{Why is shared structure inductive bias reasonable in this context?} Shared structure is a reasonable assumption in many practical applications as usually one would expect at least some similarities between treated and untreated individuals:  intuitively speaking, receiving a drug will most likely not change all biological processes related to a disease progression in a patient and attending a job training program is unlikely to neutralize all characteristics determining an individual’s salary. In medicine, for example, this has led to the well-known distinction between prognostic and predictive (effect-modifying) biomarkers \cite{ballman2015biomarker, sechidis2018distinguishing}; in our context, the strength of such prognostic information would determine the degree of shared structure. Further, assuming shared structure is compatible with explicit assumptions made in recent theoretical work on CATE meta-learners where CATE is assumed a simpler function than each of the POs \cite{kunzel2019metalearners, kennedy2020optimal, curth2020} -- which implies shared structure (i.e. a shared baseline function) between the POs. Additionally, a related (but much stronger) assumption is made in papers considering the popular semi-parametric `partially linear regression model' analyzed in e.g. \cite{robinson1988root, chernozhukov2018double}; here \textit{all} nonparametric (‘complex’) structure is shared between POs, while the treatment effect is assumed parametric (often constant).
 
 \textbf{Are there any theoretical expectations for CATE estimation performance?} All theoretical results focusing on estimating the difference between two functions that we are aware of originate in the literature on (i) CATE estimation \cite{kunzel2019metalearners, kennedy2020optimal, curth2020} or (ii) transfer learning via offset estimation \cite{wang2015generalization, du2017hypothesis}. They provide risk bounds for CATE estimation of the form $Rate_\tau + Rate_{Remainder}$, indicating that, if some remainder terms decay sufficiently fast, oracle rates for estimation of CATE can be attained. These results rely on two-stage estimation and stability conditions on the estimators, and as such are not directly applicable to our setting. Nonetheless, we hypothesize that our end-to-end learning strategies can match the performance of such estimators particularly in small sample regimes due to sharing of information between tasks (estimation of POs). It would therefore be an interesting next step to adapt theoretical results from MTL \cite{baxter2000model, maurer2016benefit} to analyze end-to-end strategies for CATE estimation, yet we consider this non-trivial as (i) MTL is concerned with the average performance over tasks, and not the performance on estimating task differences, and (ii) the number of tasks in our context is small ($T=2$). We therefore defer theoretical analysis of our approaches to future research and focus on experimentally evaluating their performance below.

\section{Experiments}
\subsection{Experimental setup}
\textbf{Simulation settings\footnote{Code to replicate all experiments is available at \url{https://github.com/AliciaCurth/CATENets}}} As ground truth treatment effects are unobserved in practice, we use semi-synthetic setups based on real covariates and simulated $\mu_w(x)$. To systematically gain insight into the relative strengths of different strategies, we consider a number of setups (A-D) comprising a total of 101 simulation settings. We provide brief descriptions below; refer to Appendix C for more detail. For setups A\&B, we use the ACIC2016 covariates ($n=4802, d=55$) of \cite{dorie2019automated} but design our own response surfaces, allowing us to introduce our own `experimental knobs' to enable structured evaluation of different approaches. We simulate response surfaces similar to  \cite{sechidis2018distinguishing} as
\begin{equation}
   \textstyle Y_i = c + \sum^d_{j=1}\beta_j (1-W_i\omega_j) X_j + \sum^d_{j=1}\sum^d_{l=1}\beta_{j,l} (1-W_i\omega_{j, l}) X_j X_l + W_i \sum^d_{j=1} \gamma_j X_j + \epsilon_i
\end{equation}
where $\beta_j, \gamma_j, \omega_j , \beta_{j,l}, \omega_{j, l} \in \{0, 1\}$ are sampled as Bernoulli random variables, and the probability of non-zero $ \beta_{j}, \beta_{j,l}$ is fixed throughout. In Setup A, we set all $\omega_{j}=\omega_{j, l}=0$ and control the complexity of $\tau(x)=\textstyle \sum \gamma_j X_j$ by varying the expected proportion $\rho$ of non-zero $\gamma_j$; thus $\mu_0(x)$ is sparser than $\mu_1(x)$. Conversely, in setup B  we set all $\gamma_j=0$ and instead induce treatment effect heterogeneity by sampling non-zero $\omega_j, \omega_{j, l}$ with probability $\rho$. These cancel some prognostic effects for the treated, such that the complexity of $\tau(x)$ increases as the complexity of $\mu_1(x)$ decreases; here $\mu_1(x)$ is thus simpler than $\mu_0(x)$. In both settings, knob $\rho$ determines the complexity of $\tau(x)$ through the number of predictive features; as $\rho$ increases, $\mu_1(x)$ becomes less similar to fixed $\mu_0(x)$. We randomly assign $n_0 \in \{200, 500, 2000\}$ to control and $\{n_1 \in \{ 200, 500, 2000\}: n_1 \leq n_0\}$ to treatment, creating different levels of sample imbalance (\textit{without} confounding), and use $500$ units for testing.

For setups C\&D, we use the IHDP benchmark ($n\!=\!747$, $d\!=\!25$), into which \cite{hill2011bayesian} introduced confounding, imbalance (18\% treated) and incomplete overlap. We  use \cite{hill2011bayesian}'s original simulation as setup C, consisting of simulated expected POs $\mu^C_0(x) = \exp((X+0.5)\beta)$ and $\mu^C_1=c + X\beta$, where $\beta$ is a randomly sampled sparse vector. Both response surfaces share the dependency on $X\beta$ albeit in different functional form; here the assumptions of additive similarity and $\tau(x)$ being simpler than $\mu_w(x)$ do \textit{not} hold and there is very strong heterogeneity.  We add setup D, in which these assumptions do hold, by slightly altering the response surfaces to $\mu^D_0(x)=\mu_0^C(x)$ and $\mu_1^D(x)=\mu_0^C(x)+\mu_1^D(x)$ (i.e. an additive treatment effect). Here, we use the 90/10 train-test splits of \cite{shalit2017estimating}'s IHDP-100 benchmark to evaluate in- and out-of-sample performance. In Appendix E, we report additional results (102 additional settings) using the original ACIC2016 response surfaces and the real-world Twins dataset.

\textbf{Models}
We compare the three approaches to TNet and TARNet as baseline indirect estimators and \cite{kennedy2020optimal}'s DR-learner, a direct two-stage estimator, which we implement using both TNet and TARNet in the first stage. We also considered  RA- \cite{curth2020}, R- \cite{nie2017quasi} and X-learner \cite{kunzel2019metalearners}, but found them to perform worse than the DR-learner on average (see appendix D). We perform further baseline comparisons, using DragonNet \cite{shi2019adapting} and SNet \cite{curth2020} in appendix D. To ensure fair comparison across all models and all experiments,  we fix hyperparameters  across all models within each experimental study and ensure that each estimator/output function has access to the same number of hidden units across all models\footnote{Due to different degrees of parameter sharing, a TNet has more parameters than TARNet and FlexTENet.} and effectively use each model ``off-the-shelf''. We discuss implementation further in Appendix C.

 \begin{figure*}[t]
    \centering
    \includegraphics[width=\textwidth]{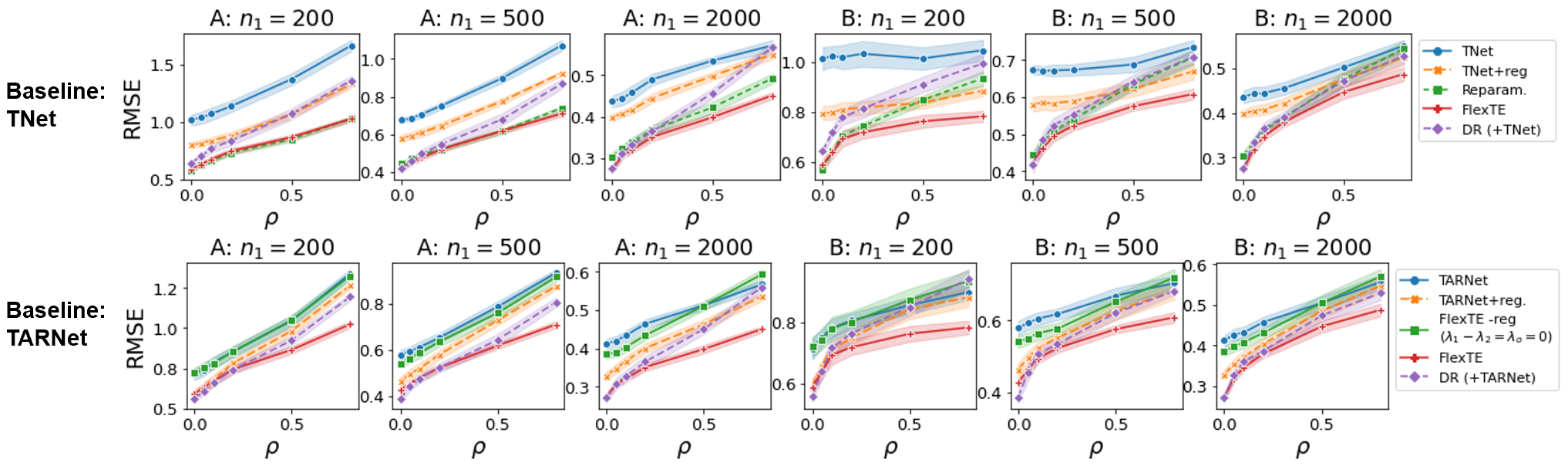}
        \vspace{-0.25in}
    \caption{RMSE of CATE estimation by $\rho$ for multiple $n_1$ for setup A and B, using TNet (top) and TARNet (bottom) as baseline, at $n_0=2000$. In each graph, as $\rho$ increases $\tau(x)$ becomes more complex because $\mu_1(x)$ becomes less (more) sparse. Avg. across 10 runs, one standard error shaded.}
    \label{fig:allRes}
          \vspace{-0.2in}  
\end{figure*}

\subsection{Results and Discussion} 
Below, we discuss our findings. We begin with setups A\&B and evaluate performance on CATE estimation with imbalanced data (where $n_0=2000$ unless stated otherwise). We perform ablations to gain insight into the relative effect of the different components of each strategy, consider the effect of smaller $n_0$ and consider performance on PO estimation. Finally, we consider setups C\&D and ACIC2016 \& Twins (Appendix E) to consolidate our findings on well-established benchmark datasets. We highlight some aspects of the results here, and present additional findings in appendix D (including varying $n_0$, further results on estimating the POs, analysing the weights of a trained FlexTENet and additional baseline comparisons).

{\textbf{Comparison to indirect learners: }\textit{All three approaches improve upon baseline indirect learners in CATE estimation with large $n_0$ (setups A\&B); regularization brings smallest gains, reparametrization works well in setup A but not B, and FlexTENet performs best on average.}} Considering Fig. \ref{fig:allRes}, we make a number of interesting observations. First, all three strategies consistently improve upon the respective baseline in almost all settings and largest improvements are made  relative to TNet with $n_1$ small. Additionally, the performance of the three approaches is closest when $\rho$ is small ($\tau(x)$ is sparse). Second, gains for regularization are most apparent with TNet for $n_1\leq 500$. Adding regularization to TARNet brings relatively smaller improvements because its baseline error is smaller. Third, for  $n_1\leq 500$, FlexTENet and reparametrization perform equivalently in setups A but not in B, which is to be expected because the latter uses an impractical parametrization for B, while FlexTENet can freely adapt to the underlying problem structure. Fourth, comparing FlexTENet and TARNet, we observe that when we use only the FlexTENet architecture without additional regularisation  ($\lambda_o=0$ and $\lambda_1=\lambda_2$), their performance is similar, while incorporating additional inductive bias, encouraging identification of predictive and prognostic effects, leads to substantial improvement of FlexTENet over TARNet across all setups.  Finally, FlexTENet performs best on average and seems to have a  particular advantage for larger $n_1$ and $\rho$, where the latter indicates that FlexTENet is not only well-equipped to handle prognostic effects, but also treatment effect heterogeneity. 

{\textbf{Comparison to multi-stage learners: }\textit{Regularization rarely outperforms the DR-learner while FlexTENet and reparametrization generally do.}} Unlike the other two approaches, regularization rarely outperforms the DR-learner, yet it often matches its performance (particularly in setup B).  Both FlexTENet and reparametrization outperform the DR-learner, yet performance of reparametrization and DR-learner seem to converge as $n_1$ increases, particularly in setup B. In Appendix D, we additionally investigate using the soft- and flexible approach as the first stage of DR- and X-learner. 

 \begin{wrapfigure}[9]{r}{0.45\textwidth}
 \vskip -0.25in
    \centering
    \includegraphics[width=0.43\textwidth]{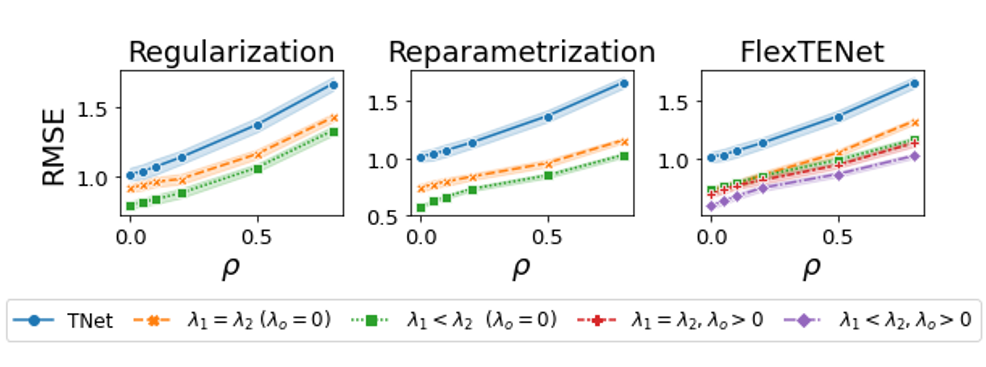}
      \vskip -0.05in
    \caption{RMSE of CATE estimation by $\rho$ for ablations ($n_1=200$, setup A). Avg. across 10 runs, one standard error shaded.}
    \label{fig:ablation}
    \vskip -0.15in
\end{wrapfigure}\textbf{Ablation study: }\textit{All components contribute to better performance. } We study ablations for setup A, $n_1=200$ (Fig. \ref{fig:ablation}) to gain insight into the relative effect of different components of each approach. We find that setting $\lambda_1<\lambda_2$, i.e. regularizing CATE more heavily than the POs, indeed led to additional (albeit smaller) improvement. For FlexTENet, orthogonal regularization alone adds more than setting $\lambda_1<\lambda_2$, yet together they lead to the greatest improvement.

\begin{wrapfigure}[8]{r}{0.45\textwidth}
 \vskip -0.23in
    \centering
    \includegraphics[width=0.43\textwidth]{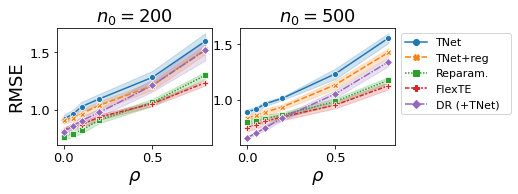}
      \vskip -0.05in
    \caption{RMSE of CATE estimation by $\rho$ for $n_0=200, 500$ ($n_1=200$, setup A). Avg. across 10 runs, one standard error shaded.}
    \label{fig:n0res}
    \vskip -0.15in
\end{wrapfigure}\textbf{Effect of $n_0$:} \textit{Improvements over baseline are less substantial for smaller $n_0$.} We investigate the effect of having a smaller set of control units $n_0$ in Fig. \ref{fig:n0res}. While all strategies continue to outperform TNet, we observe that their gains are much smaller than for $n_0=2000$ (Fig \ref{fig:allRes}). Additionally, we observe that FlexTENet appears to perform somewhat less well for small $\rho$, indicating that it may need more training data than other approaches due to its flexibility. 

\begin{wrapfigure}[10]{r}{0.47\textwidth}
    \vskip -0.15in
    \centering
    \includegraphics[width=0.45\textwidth]{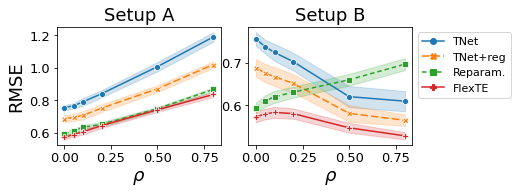}
      \vskip -0.05in
    \caption{RMSE of $\mu_1(x)$ estimation by $\rho$ for $n_1=500$ for setup A and B. Averaged across 10 runs, one standard error shaded.}
    \label{fig:PO1}
    \vskip -0.15in
\end{wrapfigure}\textbf{PO estimation}: \textit{ Most approaches also lead to improvements in PO estimation.} We consider whether the three approaches improved not only estimation of CATE but also of the POs separately in Fig. \ref{fig:PO1}. For estimation of $\mu_1(x)$ at $n_1=500$, we observe that almost all strategies improve upon TNet, but make two interesting observations: First, the performance gap between TNet and its regularized version is qualitatively smaller than the performance gap in estimating CATE (particularly in setup B), indicating that regularization could have a larger impact on improving the CATE estimate than the PO estimate. Second, the reparametrization approach is indeed unable to handle setup B where $\mu_1(x)$ becomes progressively simpler. Interestingly, even though it performs \textit{worse} at estimating $\mu_1(x)$ than TNet for large $\rho$, Fig. \ref{fig:allRes} still showed better performance at estimating $\tau(x)$ in this case -- indicating that the reparametrization approach is better targeted towards CATE estimation than PO estimation.

\begin{wraptable}[12]{r}{0.55\textwidth}
\centering
\footnotesize
\vskip -0.2in
\caption{Normalized\protect\footnotemark~  in- \& out-of-sample RMSE of CATE, setup C \& D. Avg. across 100 runs, standard error in parentheses.}\label{table:ihdp}
\vskip -0.07in
\setlength\tabcolsep{2pt}
\label{tab:ihd}
\begin{tabular}{lllll}\toprule
             & C, in      & C, out     & D, in      & D, out     \\ \midrule
TNet         & 0.32 (.01) & 0.34 (.01) & 0.29 (.01) & 0.29 (.01) \\
TNet + reg   & 0.30 (.01) & 0.32 (.01) & 0.26 (.01) & 0.26 (.01) \\
DR (+TNet)   & 0.35 (.01) & 0.37 (.01) & 0.22 (.01) & 0.22 (.01) \\
TARNet       & 0.29 (.01) & 0.31 (.01) & 0.22 (.01) & 0.23 (.01) \\
TARNet + reg & 0.28 (.01) & 0.31 (.01) & \textbf{0.20} (.01) & \textbf{0.20} (.01) \\
DR (+TARNet) & 0.33 (.01) & 0.34 (.01) & \textbf{0.20} (.01) & \textbf{0.20} (.01) \\
Reparam.     & 0.39 (.01) & 0.40 (.01) & \textbf{0.20} (.01) & \textbf{0.20} (.01) \\
FlexTENet     & \textbf{0.27} (.01) & \textbf{0.29} (.01) & 0.22 (.01) & 0.23 (.01) \\
 \bottomrule
\end{tabular}
\vskip -0.1in
\end{wraptable}\footnotetext{{Due to the exponential in $\mu_0(x)$, RMSE varies by orders of magnitude across runs and seems unsuitable to assess relative performance. We report RMSE normalized by standard deviation of observed outcomes (see appendix for details and unnormalized results). }}\textbf{Further benchmark results: }\textit{ Performance on IHDP, ACIC2016 and Twins setups reinforces findings and confirms expectations.} Even though the IHDP dataset is smaller, subject to confounding and has limited overlap, performance across setups C and D (Table \ref{table:ihdp}) largely confirms the findings previously discussed. The only major difference is induced by setup C in which $\tau(x)$ is \textit{not} simpler than each $\mu_w(x)$ separately and an additive parametrization does not lead to the easiest learning problem; as expected both reparametrization and the DR-learner perform poorly in this scenario (yet methods using shared representations are nonetheless able to exploit the shared dependence of the POs on $X\beta$). TARNet in combination with the soft approach and FlexTENet show the best average performance across the two setups, where the weaker performance of FlexTENet on setup D is in line with previous findings for small $n_0$ and simple $\tau(x)$. The results on the ACIC2016 simulations and the real-world Twins dataset presented in appendix E further confirm the relative performance of the different methods we observed throughout.

\section{Conclusion} We found that altering the inductive biases in end-to-end CATE estimators can lead to performance increases and match, or improve upon, the finite sample performance of multi-stage learners targeting CATE directly. We observed that all our approaches were particularly useful when only one treatment group had abundant samples, a situation arising naturally in practice when a new treatment is introduced. We found that strategies which change the model architecture -- reparametrization and FlexTENet -- led to the largest improvements: The former seems to be the best choice in smaller samples when treatment effects are additive and $\mu_0(x)$ is simpler than $\mu_1(x)$, while FlexTENet showed the best average performance due to its flexibility and handled high heterogeneity well, but required more training data when $\tau(x)$ is very simple. Additionally, we also found that the simple strategy of changing the regularizer leads to notable performance increases, an insight that could directly be incorporated into many existing methods with treatment-specific parameters. 

Here, we limited our attention to a single ML method (NNs) to \textit{isolate} the properties of different approaches. Therefore, an interesting next step would be to consolidate our findings by applying our approaches using different underlying methods. Finally, as all approaches rely on the ability to manipulate loss (or likelihood) functions, it would be interesting to consider how to equip methods that do not have easily manipulable loss functions (e.g. random forests) with similar inductive biases.

\acksection
We thank anonymous reviewers as well as members of the vanderschaar-lab for many insightful comment and suggestions. AC gratefully acknowledges funding from AstraZeneca.

\bibliographystyle{unsrt}
\bibliography{main}

 \newpage
 \appendix
 \section*{Appendix}
This appendix is organized as follows: We first present an additional literature review (Section A) and then discuss additional details of the proposed approaches (Section B).We then provide further experimental details (Section C), discuss additional results on Setups A-D (Section D) and finally present results on additional benchmark datasets (Section E). We include the NeurIPS checklist in Section F.

\section{Additional Literature Review}
Here, we present a detailed overview of existing model-agnostic ``meta-learner'' strategies for CATE estimation which can be implemented using \textit{any} ML method, a notion originally introduced in \cite{kunzel2019metalearners} and expanded on in \cite{nie2017quasi, kennedy2020optimal, curth2020}. As in the main text, we distinguish between indirect and direct estimators for CATE, and will finally briefly discuss ML-based strategies that do \textit{not} fall in the meta-learner class because they rely on a specific ML-method. 

\textbf{Indirect Estimators}
The S- and T-learner discussed in \cite{kunzel2019metalearners} are two model-agnostic learning strategies that estimate CATE \textit{indirectly}. The S-learner fits a \textbf{s}ingle regression model $\hat{\mu}(x, w)$ by concatenating the covariate vector $X$ and the treatment indicator $W$ into $X'$ and then regressing $Y$ on $X'$, providing a final CATE estimate indirectly as $\hat{\tau}(x)= \hat{\mu}(x, 1) -  \hat{\mu}(x, 0)$. The T-learner fits \textbf{t}wo regression models (one $\hat{\mu}_w(x)=\mathbbm{E}[Y|W=w, X=x]$ for each treatment group) \textit{separately} using only observations for which $W=w$, and provides a final CATE estimate as $\hat{\tau}(x)= \hat{\mu}_1(x) -  \hat{\mu}_0(x)$. 

\textbf{Multi-stage Direct Estimators}
A number of meta-learners have been proposed recently which target CATE \textit{directly} through a multi-stage estimation procedure. We will first discuss four learning strategies that rely on \textit{pseudo-outcome} regression, and will then discuss \cite{nie2017quasi}'s R-learner, which uses a loss-based approach. 

We follow the exposition in \cite{curth2020} distinguishing between three classes of pseudo-outcome regression-based meta-learners, which use a first stage to obtain estimates $\hat{\eta}$ of (a subset of) the nuisance parameters $\eta=(\mu_0(x), \mu_1(x), \pi(x))$. Pseudo-outcome regression then proceeds by obtaining an estimate of $\hat{\tau}(x)$ by regressing a pseudo-outcome $\tilde{Y}_{\hat{\eta}}$ (based on nuisance estimates $\hat{\eta}$) on $X$ directly. For all considered pseudo-outcomes it holds that $\mathbbm{E}_{\mathbbm{P}}[\tilde{Y}_{\eta}|X=x]=\tau(x)$  -- they are unbiased for CATE when $\eta$ is known. Inspired by the well-known estimation strategies for the \textit{average} treatment effect (ATE), there are three straightforward strategies for doing so:

(1) Regression Adjustment (RA): The RA-learner \cite{curth2020} uses a regression-adjusted pseudo-outcome 
\begin{equation}
\tilde{Y}_{RA, \hat{\eta}} = W(Y-\hat{\mu}_0(X)) + (1-W)(\hat{\mu}_1(X) - Y)
\end{equation}
in the second stage. The X-learner proposed in \cite{kunzel2019metalearners} is a variant of this estimator: Instead of performing one regression in the second stage, they perform two separate regressions for each term in the sum, leading to two CATE estimators $\hat{\tau}_1(x)$ and $\hat{\tau}_0(x)$ that are then combined into a final estimate using $\hat{\tau}(x) = g(x) \hat{\tau}_1(x) + (1 - g(x)) \hat{\tau}_0(x)$ for some weighting function $g(x)$. 

(2) Propensity Weighting (PW):  The PW-learner uses a pseudo-outcome based on the Horvitz-Thompson transformation \cite{horvitz1952generalization}
\begin{equation}
\tilde{Y}_{PW, \hat{\eta}} = \left(\frac{W}{\hat{\pi}(X)}- \frac{1-W}{1-\hat{\pi}(X)}\right)Y 
\end{equation}

(3) Doubly Robust (DR): \cite{kennedy2020optimal}'s DR-learner has pseudo-outcome
\begin{equation}
\begin{split}
\tilde{Y}_{DR, \hat{\eta}} = \left(\frac{W}{\hat{\pi}(X)}- \frac{(1-W)}{1-\hat{\pi}(X)}\right) Y + \left[\left(1 - \frac{W}{\hat{\pi}(X)}\right) \hat{\mu}_1(x)-\left(1 - \frac{1-W}{1-\hat{\pi}(X)}\right)\hat{\mu}_0(X)\right]
\end{split}
\end{equation}
which is based on the doubly-robust AIPW estimator \cite{robins1995semiparametric} and is unbiased if either propensity score \textit{or} outcome regressions are correctly specified. 

For two-stage estimators targeting the treatment effect directly using pseudo-outcomes, it can be shown that $\epsilon_{sq}(\hat{\tau}(x)) \leq {\epsilon_{sq}(\hat{\tau}_{\eta}(x))}+ {R^2_{\hat{\eta}}(x)}$ if appropriate sample splitting is used and the used estimator fulfills some stability condition \cite{kennedy2020optimal}. Here, ${\epsilon_{sq}(\hat{\tau}_{\eta}(x))}$ converges at the oracle rate, and $R^2_{\hat{\eta}}$ is a learner-specific remainder term. Two-stage learners thus converge at oracle rates if the remainder term decays sufficiently fast, which is faster than indirect learners if $\tau(x)$ is simple. Due to the double robustness property, the remainder term of the DR-learner always converges faster than the other two learners, and it is in general unlikely that RA-learner can attain the oracle rate. For a comprehensive overview of convergence of the different pseudo-outcome-based meta-learners, refer to \cite{curth2020}.

Finally, \cite{nie2017quasi} also propose a two-stage algorithm that estimates CATE directly but is based on a loss-based strategy instead of pseudo-outcome regression. The R-learner is based on \cite{robinson1988root} approach for semiparametric regression, and uses orthogonalization with respect to the nuisance functions $\pi(x)$ and $\mu(x)=\mathbbm{E}[Y|X=x]$ (the unconditional outcome expectation). Their first stage obtains estimates $\hat{\pi}(x)$ and $\hat{\mu}(x)$, which are then used in a second stage estimating $\tau(x)$ directly based on the following loss:
\begin{equation}
  \arg \min_\tau  \sum^n_{i=1}\left[\{Y_i - \hat{\mu}(X_i)\} - \{W_i - \hat{\pi}(X_i)\}\tau(X_i)\right]^2 + \mathcal{R}(\tau(\cdot))
\end{equation}
Like the DR-learner, this learner is doubly robust due to orthogonalization with respect to both outcome and propensity score estimate, and also arises from the perspective of `orthogonal statistical learning' \cite{foster2019orthogonal}. 

For theoretical guarantees to hold for any of the two-stage learners, first and second stage have to be performed on separate folds of the data, either by splitting the data or by using cross-fitting \cite{chernozhukov2018double}. In our experiments, we do use the full data for both stages as we found that sample splitting/cross-fitting deteriorates performance, most likely due to the small sample sizes. 

\textbf{Model-specific CATE estimators}
Many methods proposed in related work do not fall within the meta-learner class because they rely on the properties of a \textit{specific} ML method. Some methods rely on a strategy that can be seen as a hybrid between \cite{kunzel2019metalearners}'s S- and T-learner, sharing \textit{some} information between regression tasks but not all;  this includes most multi-task approaches (\cite{shalit2017estimating} and extensions, but also e.g. \cite{alaa2017deep, alaa2018limits}). GANITE \cite{yoon2018ganite}, another popular strategy, relies on GANs to learn counterfactual distributions instead of conditional expected values. Additionally, as discussed in the main text, \cite{imai2013estimating}'s LASSO for CATE estimation and \cite{hahn2017bayesian}'s Bayesian Causal Forest rely a the reparametrization strategy (while \cite{hill2011bayesian}'s popular BART-based CATE estimator is a simple S-learner). Finally, \cite{athey2019generalized}'s Causal Forest relies on a local moment equation inspired by the Robinson transformation, and could hence be seen as a forest-based version of the R-learner.

\section{Additional discussion of proposed approaches}

\subsection{Hyperparameter tuning}
It is a well-known problem in the CATE estimation literature that model selection is nontrivial due to the absence of counterfactuals in practice. The testable implications of the shared structure bias, as encoded by hyperparameters such as $\lambda_2$, however, are different for (i) the PO estimation and (ii) the CATE estimation problems, which is a feature that we would suggest to exploit in choosing hyperparameter settings. That is, while the usefulness of the inductive biases for estimation \textit{of CATE} cannot easily be verified, their usefulness for estimating \textit{the POs} can be verified through cross-validation on held-out factual observations.

Unfortunately, good performance on estimation of the POs is not sufficient. To see this, note that the tuples of PO estimates $(\mu_0(x)+e(x), \mu_1(x)+e(x))$ and  $(\mu_0(x)+e(x), \mu_1(x)-e(x))$  [where $e(x)$ is an error and $\mu_w(x)$ is the ground truth] will in expectation give exactly the same MSE when evaluated using held-out factual observations. Yet, the first tuple will make no error in estimating CATE, while the second tuple makes an error of $2e(x)$. This highlights that error can either compound or cancel across the POs, therefore making it possible that a hyperparameter setting resulting in better fit on the POs also results in worse fit on CATE, and vice versa. Because an estimators’ performance on estimating CATE remains unobservable in practice, a good inductive bias is needed to choose between models that have equal predictive performance on the POs. As we discuss in the paper, we consider preferring shared structure (or a simple CATE) a reasonable choice for this in many practical applications.

\paragraph{Suggested approach for setting hyperparameters.} As a simple heuristic to set hyperparameters such as $\lambda_2$ in practice, we would thus recommend the following scheme that trades off between imposing an assumption (“CATE is most likely simple”) and factual performance:

\begin{enumerate}
    \item Start with $\lambda_2$ small, and keep increasing it while held-out predictive performance does not decrease.
\item Set $\lambda_2$ to its largest value for which predictive performance did not deteriorate.
\end{enumerate}
In a sufficiently flexible (overparameterized) model class in which multiple PO estimators induce the same empirical performance, such a scheme allows to pick the hyperparameter setting resulting in the least complex CATE while remaining compatible with factual observations.

\begin{figure}[!h]
    \centering
    \includegraphics[width=\textwidth]{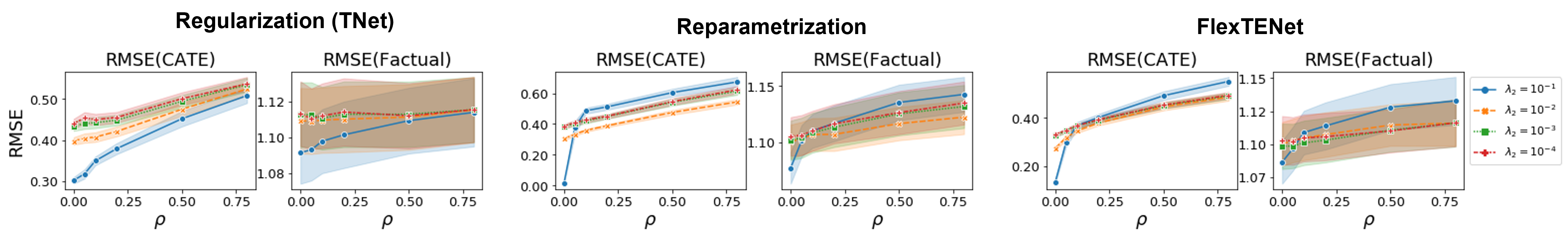}
    \vspace{-0.05in}
    \caption{Effect of $\lambda_2$ on RMSE of estimating CATE and factual RMSE (test set), for regularization (left), reparametrization (middle) and flexible (right) approach, by $\rho$ in Setup B at $n_0=n_1=2000$. Avg. across 10 runs, one standard error shaded.}
    \label{fig:hyperparams}
\end{figure}

\paragraph{Illustrative results} In Fig. \ref{fig:hyperparams} we present illustrative results on Setup B with $n_0=n_1=2000$, and observe that following our heuristic of increasing $\lambda_2$ until factual performance deteriorates would almost always lead to choosing the best hyperparameter setting; for both hard and flexible approach this suggests a switch from $\lambda=10^{-1}$ to $\lambda_2=10^{-2}$ as $\rho$ increases\footnote{Note that, as we discuss in section C.2, we fixed all hyperparameters throughout all experiments as tuning would have been computationally prohibitive given the amount of experimental settings and models considered. Throughout, we used $\lambda_1=10^{-4}$ and $\lambda_2 = 100 \lambda_1= 10^{-2}$ to induce a substantial difference between the two.}. Further, we note that performance in terms of factual prediction is indeed often much closer than CATE estimation performance \footnote{Note also that for factual evaluation observations have random normal noise with $\sigma=1$, which explains why $RMSE(factual)>1$ while $RMSE(CATE)<1$.}.

\subsection{FlexTENet as a generalization of existing architectures}
 \begin{table}[ht]
\vskip -0.1in
\footnotesize
\centering
\caption{Existing architectures for CATE estimation arise as special cases of FlexTENet}
\label{tab:struc}
\vskip 0.1in
\setlength\tabcolsep{2pt}
\begin{tabular}{@{}lccc@{}}
\toprule
Method & $(n_{r,s},n_{r,0}, n_{r,1})$ & $(n_{h,s},n_{h,0}, n_{h,1})$ & Communicat.\\ &&&subspaces \\ \midrule
\rowcolor{gray!20} FlexTENet & $(n_{r,s},n_{r,0}, n_{r,1})$ & $(n_{h,s},n_{h,0}, n_{h,1})$ & Yes \\
TARNet & $(n_r, 0, 0)$ & $(0, n_h, n_h)$ & No \\
TNet & $(0, n_r, n_r)$ & $(0, n_h, n_h)$ & No \\
SNet & $(n_{r,s},n_{r,0}, n_{r,1})$ & $(0, n_h, n_h)$ & No \\
Reparam. & $(n_r, 0, n_r)$ & $(n_h, 0, n_h)$ & No \\
 \bottomrule
\end{tabular}
\vskip -0.1in
\end{table}
 As summarized in Table \ref{tab:struc}, existing architectures for CATE estimation arise as special cases of FlexTENet (all of which heavily restrict the flow of information within the network): As we reduce the width of all shared layers while increasing the width of private layers, FlexTENet approaches a TNet. Conversely, if we increase the width of the bottom shared layers and do the reverse for the top layers, FlexTENet becomes TARNet. Relative to TARNet, this architecture thus not only incorporates inductive bias towards shared behavior in the output heads, but also explicitly anticipates the existence of purely predictive features by allowing for both shared and PO-specific features instead of enforcing one joint representation (which might erroneously discard features that are relevant only for one of the POs). Therefore, FlexTENet also generalizes the SNet class discussed in \cite{curth2020}, which includes PO-specific feature spaces\footnote{The general SNet specification in \cite{curth2020} also includes propensity estimators as additional output heads, which could be added to FlexTENet if needed.}. Finally, FlexTENet without private subspaces for $\mu_0(x)$ and without communication between shared and private subspaces is equivalent to the reparametrization approach. Relative to existing strategies, we expect that such a general architecture should perform well \textit{on average}, given that it not only encompasses them, but also allows for more general forms of shared structure which other architectures cannot exploit.

\subsection{Pseudocode of a FlexTENet forward pass}
\begin{algorithm}[h]
    \SetAlgoLined
    \SetKwInOut{Input}{Input}
    \Input{Testing data X\\Trained FlexTENet \texttt{flex}}
    \For{i $\leftarrow$ \text{1:\texttt{flex.n\_layers}}}{
    \If{i==1}{
        \texttt{x\_shared} = \texttt{flex.shared\_layers[i]}(X)\\
        \texttt{x\_po0} = \texttt{flex.po0\_layers[i]}(X)\\
        \texttt{x\_po1} = \texttt{flex.po1\_layers[i]}(X)
    }\Else{
        \texttt{x\_po0} = \texttt{flex.po0\_layers[i]}(\texttt{Concatenate}(\texttt{x\_shared},\texttt{x\_po0})\\
        \texttt{x\_po1} = \texttt{flex.po1\_layers[i]}(\texttt{Concatenate}(\texttt{x\_shared},\texttt{x\_po1})\\
        \texttt{x\_shared} = \texttt{flex.shared\_layers[i]}(\texttt{x\_shared})\\
    }
    }
    \If{\text{\texttt{flex.binary\_y}}}{
    \texttt{y0\_hat} = \texttt{Sigmoid}(\texttt{x\_shared}+\texttt{x\_po0})\\
    \texttt{y1\_hat} = \texttt{Sigmoid}(\texttt{x\_shared}+\texttt{x\_po1})
    }\Else{
    \texttt{y0\_hat} = \texttt{x\_shared}+\texttt{x\_po0}\\
    \texttt{y1\_hat} = \texttt{x\_shared}+\texttt{x\_po1}
    }
    \Return{\texttt{y0\_hat},   \texttt{y1\_hat}}
\label{alg:forward}
\caption{FlexTENet forward pass.}
\end{algorithm}

\section{Experimental details}
\subsection{Simulation details}
We considered setups A-D in the main text, as they allowed us to evaluate the performance of the different approaches along axes that are most relevant to the problem we are trying to solve. Unlike much related work (e.g. \cite{shalit2017estimating, hassanpour2020learning, assaad2020counterfactual}) we therefore did not focus on varying the level of confounding, but instead considered (i) the level of alignment between the PO functions and (ii) differences in data availability between the two treatment groups. While (i) directly determines how applicable/useful the underlying inductive biases are, (ii) is of interest because we wanted to test whether a large control group allows to distill prognostic effects better, such that less treatment data is needed to determine predictive effects. 

Major differences across setups include that: 
\begin{itemize}[topsep={0pt},itemsep={0pt}]
    \item in setups A, B, and D, the treatment effects are additive, while setup C inherits a non-additive treatment effect from \cite{hill2011bayesian}'s IHDP simulation
\item in setups A \& B there is no confounding as treatment assignment is random, while in C \& D, there is confounding and incomplete overlap 
\item in setups A \& B, the response surface includes only polynomial terms of the covariates (linear, squares and first-order interactions) while in C \& D the baseline outcome exponentiates a linear predictor 
\item in setups A \& D, $\mu_0(x)$ is simpler than $\mu_1(x)$, while the reverse is true in setups B \& C.  
\end{itemize}

Note that, while Setups C \& D corresponded to one simulation setting each, setups A \& B as presented in Fig. 3 in the main text corresponded to 33 simulation settings (as $\rho$ and $n_1$ vary). By further varying $n_0$, we consider another 66 settings in section \ref{n0appendix}. Below, we give further insight into the data and data-generating processes (DGPs) we used. 
\subsubsection{Setups A and B}
We use the data from the Collaborative Perinatal Project provided\footnote{We retrieve the data from \url{https://jenniferhill7.wixsite.com/acic-2016/competition}} in the first Atlantic Causal Inference Competition (ACIC2016) \cite{dorie2019automated} for our first set of experiments. The original dataset has $d=58$ covariates, of which we exclude the $3$ categorical ones. Of the remaining $55$ covariates, $5$ are binary, $27$ are count data and $23$ are continuous. We process all covariates according to the transformations used in the competition\footnote{We use the code available at \url{https://github.com/vdorie/aciccomp/blob/master/2016/R/transformInput.R}}, which transforms count into binary covariates and standardizes continuous variables. We use the transformed data for the simulations and as input to all models.

As discussed in the main text, we simulate response surfaces in setup A according to
\begin{equation}
    Y_i = c + \sum^d_{j=1}\beta_j X_j + \sum^d_{j=1}\sum^d_{l=1}\beta_{j,l} X_j X_l + W_i \sum^d_{j=1} \gamma_j X_j + \epsilon_i
\end{equation}
where $\epsilon_i \sim N(0, 1)$, $\beta_j\sim \mathcal{B}(0.6)$ and $\gamma_j\sim \mathcal{B}(\rho)$. We include squared terms of all continuous covariates and additionally include each variable randomly into one interaction term, for both of which we then simulate coefficient $\beta_{j, l}\sim \mathcal{B}(0.3)$. We chose for each coefficient to be binary to avoid large variances in the scale of POs and CATE across different runs of a simulation, such that RMSE remains comparable across runs. 

For setup B, we instead use $\gamma_j=0$ and simulate 
\begin{equation}
    Y_i = c + \sum^d_{j=1}\beta_j(1-W_i\omega_j) X_j + \\\sum^d_{j=1}\sum^d_{l=1}\beta_{j,l} (1-W_i\omega_{j, l})X_j X_l + \epsilon_i
\end{equation}
where only $\omega_j, \omega_{j, l} \sim \mathcal{B}(\rho)$ differ from above. While in setup A non-zero $\gamma_j$  induce treatment effect heterogeneity as $\mu_1(x)$ has \textit{more} terms than $\mu_0(x)$, in setup B non-zero $\omega_j$ induce treatment effect heterogeneity, giving $\mu_1(x)$ \textit{less} terms than $\mu_0(x)$. 

\subsubsection{Setups C and D (IHDP)}
For setups C \& D, we build on the Infant Health and Development Program (IHDP) benchmark used in \cite{shalit2017estimating} and extensions, created by \cite{hill2011bayesian}. The underlying dataset belongs to a real randomized experiment targeting premature infants with low birth weight with an intervention, containing 25 covariates (6 continuous, 19 binary) capturing aspects related to children and their mothers. The benchmark dataset was created by excluding a non-random proportion of treated individuals (those with nonwhite mothers). The final dataset consists of 747 observations (139 treated, 608 control), and overlap is not satisfied (as $\pi(x)$ is not necessarily non-zero for all observations in the control group). While the covariate data is real, the outcomes in our setup ``C'' are simulated according to setup ``B" described in \cite{hill2011bayesian}, which satisfies $Y(0) \sim \mathcal{N}(exp((X+W)\beta), 1)$ and $Y(1) \sim \mathcal{N}(X\beta - \omega, 1)$ with $W$ an offset matrix, $\omega$ is set such that the average treatment effect on the treated is equal to 4, and the coefficient $\beta$ has entries in $(0, 0.1, 0.2, 0.3, 0.4)$, where each entry is independently sampled with probabilities $(0.6, 0.1, 0.1, 0.1, 0.1)$. We use the 100 repetitions of the simulation provided by \cite{shalit2017estimating}\footnote{Available at \url{https://www.fredjo.com/}}. For our setup ``D'' we change only the response surface of the treated to $Y(1) \sim \mathcal{N}(exp((X+W)\beta) + X\beta - \omega, 1)$.

\subsection{Implementation details}\label{sec:implementation}
In our implementations\footnote{Code to replicate all experiments is available at \url{https://github.com/AliciaCurth/CATENets}}, we use components similar to those used in \cite{shalit2017estimating} for all networks. In particular, we use dense layers with exponential linear units (ELU) as nonlinear activation functions. We train with Adam \cite{kingma2014adam}, minibatches of size 100, and use early stopping based on a 30\% validation split. As stated in the main text, we fixed equivalent hyperparameters across all methods within any experiments to not conflate hyperparameter tuning with the value of the different strategies.  We set $n_r=200$ and $n_h=100$ throughout, and use $d_r=1$ and $d_h=1$ for setups A \& B and $d_r=2$ and $d_h=2$ for setups C \& D (excluding one additional dense output layer) for all estimators -- i.e. for TARNet, but also for each other function (e.g. the second stage of a two-step learner is parameterized by $d_r$ and $d_h$ layers of $n_r$ and $n_h$ units each). Note that TNet and TARNet with similarity regularization share the identical architecture with their `vanilla' counterparts, and differ only in the regularization term and in that we initialise the $\Theta_{h_w}$ weights with the same random initialisation for both heads. For FlexTENet we set $n_{k, p}=n_{k, s}=\frac{1}{2}n_k$ for $k \in \{r, h\}$. Throughout, we use $\lambda_1=0.0001$, $\lambda_2 = 100 \lambda_1$ (to induce a substantial difference) and $\lambda_o=0.1$, where we chose the magnitude of $\lambda_o$ by testing it on toy data. Further, we use all two-step learners \textit{without} sample splitting or cross-fitting which we found to deteriorate performance, particularly in the smaller sample sizes. 

All models were implemented in our own python codebase, using jax \cite{jax2018github}. All experiments were conducted using Python 3.8 on Ubuntu 20.04 OS with a Intel i7-8550U CPU with 4 cores. Creating the results in Figure 3 and Table 1 (main text) took about 12h each.

\section{Additional results (setups A-D)}
Below, we present additional results. First, we consider additional baselines: we compare the performance of different two-stage learners across setups A \& B (D.1) and consider further indirect learners as baselines (DragonNet and SNet, D.2). Then, we consider the effect of $n_0$ in setups A \& B (D.3), present additional results on PO estimation (D.4), and then move to analyzing the learned weights of a FlexTENet (D.5). We also consider the effect of using our approaches as first-stage (nuisance) estimators for two-step learners (D.6). Finally, we  discuss the necessity of scaling the results in setups C \& D (D.7)

\subsection{Comparison of two-stage learners across setups A and B}
\begin{figure}[!h]
    \centering
    \includegraphics[width=\textwidth]{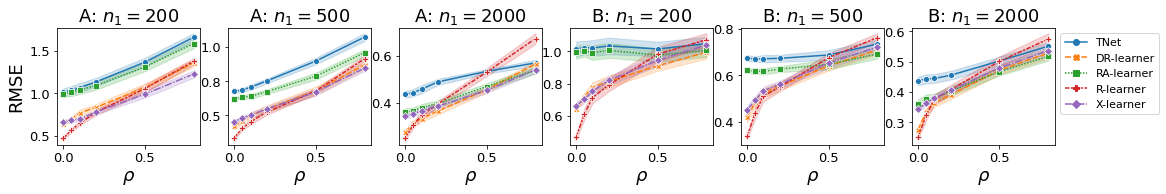}
    \vspace{-0.2in}
    \caption{RMSE of CATE estimation for two-step learners using TNet as baseline by $\rho$ for different $n_1$ with $n_0=2000$ for setup A ($\mu_0(x)$ is simpler) and B ($\mu_1(x)$ is simpler). Recall that, in each graph, as $\rho$ increases $\tau(x)$ becomes more complex because $\mu_1(x)$ becomes less (more) sparse. Avg. across 10 runs, one standard error shaded.}
    \label{fig:twostep}
\end{figure}

When comparing the performance of DR-learner, R-learner, RA-learner and X-learner (Fig \ref{fig:twostep}), we observe that the DR-learner shows best performance \textit{on average} -- which is why we used it as a baseline in the main text. Nonetheless, the R-learner can outperform it for small $\rho$, while the X-learner can outperform it for large $\rho$.

\subsection{Experiments with additional indirect learners}
\begin{figure}[!h]
    \centering
    \includegraphics[width=\textwidth]{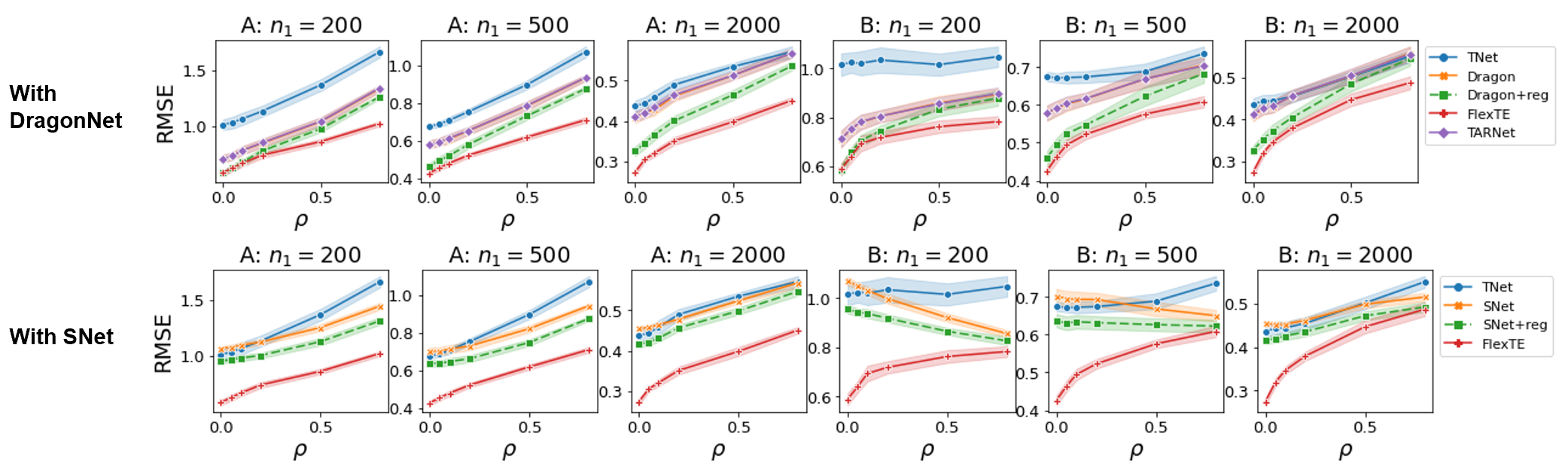}
    \vspace{-0.2in}
    \caption{RMSE of CATE estimation using DragonNet (top) and SNet (bottom) as additional baselines, by $\rho$ for multiple $n_1$ at $n_0=2000$, for setup A and B. Avg. across 10 runs, one standard error shaded.}
    \label{fig:baseline}
\end{figure}

In Fig. \ref{fig:baseline}, we present additional results using DragonNet \cite{shi2019adapting} and SNet \cite{curth2020} on setups A \& B. As there is no confounding in these setups, the propensity head of DragonNet does not contribute to performance, such that TARNet and DragonNet perform virtually equivalently across all settings, and the soft approach improves also the performance of DragonNet.

For SNet, to facilitate comparison, we consider a variant of the original formulation in \cite{curth2020} without propensity head, such that SNet reduces to having only 3 feature spaces - a shared and two PO-specific feature spaces just like FlexTENet. As such, SNet and FlexTENet differ mainly in their output heads. Further, we implement SNet using the same orthogonal regularizer as FlexTENet to allow for fair comparison (this differs from the original orthogonal regularizer used in \cite{curth2020}, which induces more sparsity as it relied on a l1-norm). We observe that the soft approach improves also the performance of SNet, and that FlexTENet substantially outperforms SNet despite their similarities; highlighting that hard-sharing in the output heads is useful.

\begin{wraptable}[14]{r}{0.7\textwidth}
\centering
\footnotesize
\vskip -0.2in
\caption{Normalized  in- \& out-of-sample RMSE of CATE estimation for additional benchmarks and selected methods, setup C \& D. Avg. across 100 runs, standard error in parentheses.}\label{table:ihdp_extra}
\vskip -0.07in
\setlength\tabcolsep{2pt}
\begin{tabular}{lllll}\toprule
             & C, in      & C, out     & D, in      & D, out     \\ \midrule
TNet         & 0.320 (.008) & 0.337 (.008) & 0.290 (.007) & 0.290 (.008) \\
TNet + reg   & 0.301 (.008) & 0.324 (.008) & 0.260 (.005) & 0.262 (.006) \\
TARNet       & 0.294 (.008) & 0.315 (.008) & 0.225 (.007) & 0.226 (.007) \\
TARNet + reg & 0.285 (.008) & 0.306 (.008) & 0.205 (.006) & 0.205 (.006) \\
FlexTENet     & \textbf{0.268} (.009) & \textbf{0.293} (.009) & 0.224 (.005) & 0.230 (.006) \\ 
DragonNet & 0.289 (0.008) & 0.310 (0.008) &0.222 (0.006) & 0.223 (0.007) \\
DragonNet + reg & 0.282 (0.008) & 0.304 (0.008) & \textbf{0.203} (0.006)& \textbf{0.203} (0.006) \\
SNet & 0.327 (0.014) & 0.356 (0.013) & 0.261 (0.008) & 0.266 (0.009) \\
SNet + reg & 0.316 (0.013) & 0.345 (0.012) & 0.252 (0.008) & 0.258 (0.009) \\
 \bottomrule
\end{tabular}
\vskip -0.1in
\end{wraptable}
Results for Setups C and D are presented in table \ref{table:ihdp_extra}. We observe that DragonNet once more performs very similar to TARNet; here it does perform slightly better, possibly due to the presence of confounding in this dataset. However, as $\mu_w(x)$ and $\pi(x)$ are not well-aligned, the gains are marginal. SNet performs poorly overall, which is to be expected as there are no $\mu_w(x)$-specific features in these setups. Further, both baselines benefit from the addition of the soft approach.

\subsection{The effect of $n_0$ in setups A and B}\label{n0appendix}
\begin{figure}[!h]
    \centering
    \includegraphics[width=\textwidth]{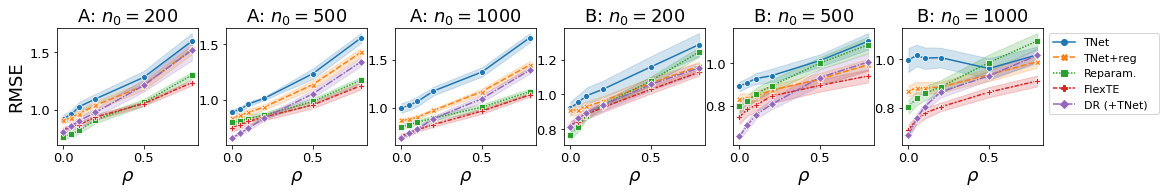}
    \vspace{-0.2in}
    \caption{RMSE of CATE estimation by $\rho$ for $n_0=200, 500, 1000$ for setup A and B with $n_1=200$ using TNet as baseline. Avg. across 10 runs, one standard error shaded.}
    \label{fig:n0abl}
\end{figure}

In Fig. \ref{fig:n0abl} we investigate the effect of having a smaller number of control samples, $n_0=200, 500, 1000$ instead of $n_0=2000$ as in the main text, at $n_1=200$. We observe that in setup A, increasing $n_0$ leads to convergence of performance of both regularization and DR-learner as well as of FlexTENet and reparametrization. Interestingly, in setup B, we observe divergence of the different methods.

\begin{figure}[!h]
    \centering
    \includegraphics[width=\textwidth]{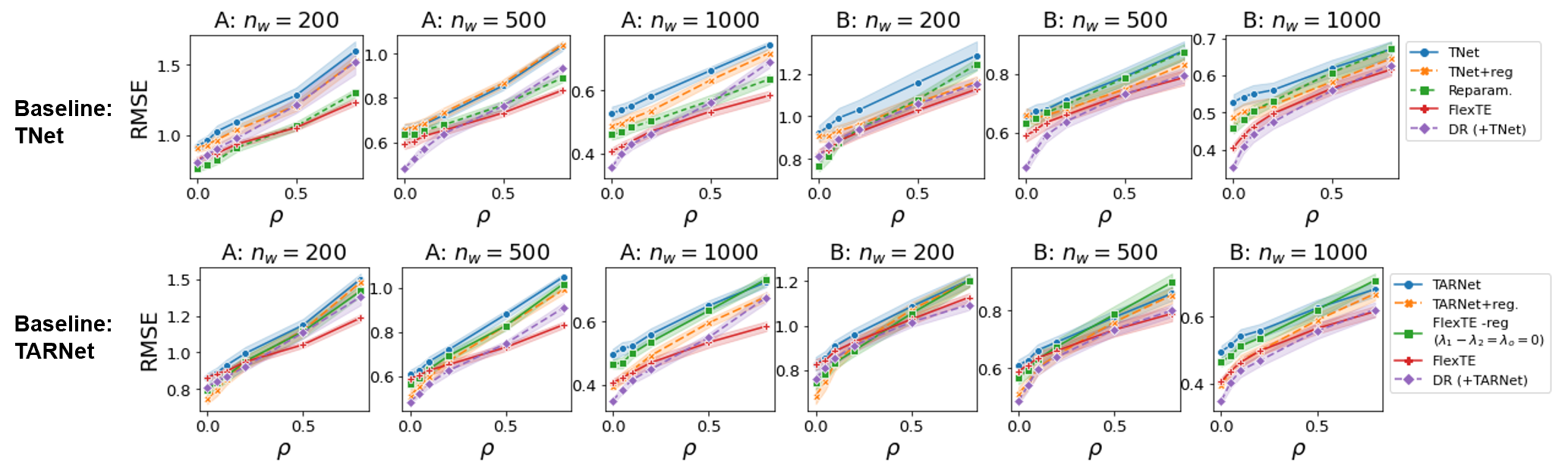}
        \vspace{-0.25in}
    \caption{RMSE of CATE estimation by $\rho$ for $n_0= n_1=200, 500, 1000$ for setup A and B, using TNet (top row) and TARNet (bottom row) as baseline. Avg. across 10 runs, one standard error shaded.}
    \label{fig:allbalanced}
\end{figure}

Additionally, we consider increasing the number of observations in treatment and control group equally, i.e. $n_0=n_1$, for $n_w\in\{200, 500, 1000\}$ in Fig. \ref{fig:allbalanced}  ($n_0=n_1=2000$ is included in the results in the main text). As briefly discussed in the main text, the gain of using each approach is much smaller in the balanced than in the imbalanced setting, but the conclusions regarding the relative performance of each approach remain largely the same. Most salient is the strong performance of the DR-learner for small $\rho$ (often outperforming all other methods) and in setup B throughout (often matching the performance of FlexTENet for large $\rho$).

\subsection{Additional results for PO estimation}
\begin{figure}[!h]
    \centering
    \includegraphics[width=0.7\textwidth]{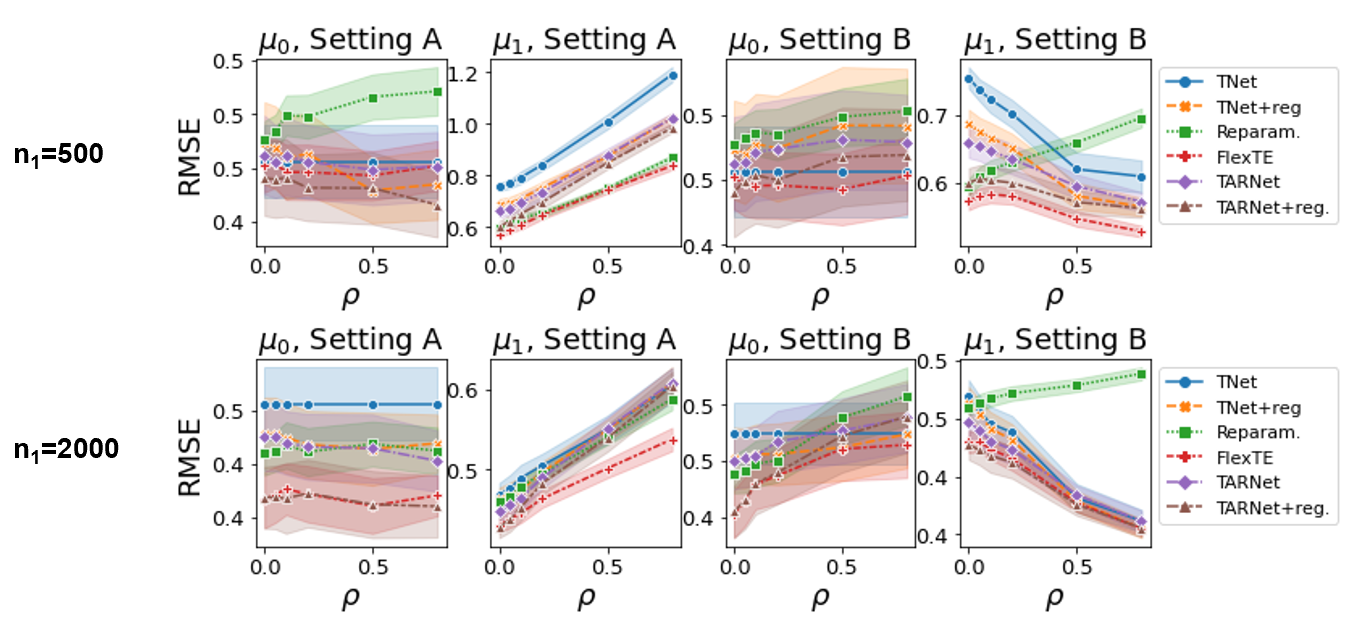}
    \vspace{-0.1in}
    \caption{RMSE of $\mu_w(x)$ estimation by $\rho$ for $n_1=500$ (top) and $n_1=2000$ (bottom) for setup A and B at $n_0=2000$. Avg. across 10 runs, one standard error shaded.}
    \label{fig:poextra}
\end{figure}

In Fig. \ref{fig:poextra}, we observe that the impact of all approaches on $\mu_0(x)$ estimation is negligible for small $n_1$ (at $n_0=2000$), but that there are some improvements for $n_1=2000$. Further, the inability of reparametrization to handle setup B is even more apparent for $n_1=2000$.

 \begin{figure}[!h]
    \centering
    \includegraphics[width=\textwidth]{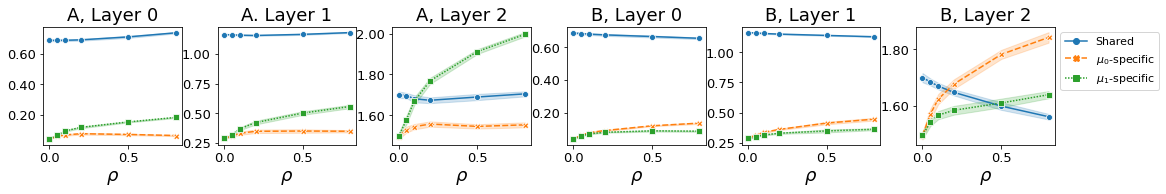}
    \vspace{-0.2in}
    \caption{Average L2-norm of the weights of each hidden unit for each layer and subspace of the FlexTENet  by $\rho$, for setup A (left) and B (right) at $n_0=n_1=2000$. Avg. across 10 runs, one standard error shaded.}
    \label{fig:weights}
\end{figure}

\subsection{Analysis of FlexTENet weights}
In Fig. \ref{fig:weights} we analyze what the FlexTENet learns by considering the average L2-norm of the weights of each hidden unit for each layer and subspace. We observe that in the lower layers, most weight is on the shared component for all $\rho$, and that the weight on the $\mu_1(x)$-specific ($\mu_0(x)$-specific) component increases with increasing $\rho$ in setup A (B), as expected. Further, while in the output layer the weight on the shared layer remains approximately constant with $\rho$ in setup A, it decreases in setup B where the shared component becomes sparser as $\rho$ increases.

\subsection{Using the the approaches as improved first-stage estimators for two-stage learners}
\begin{figure}[!h]
    \centering
    \includegraphics[width=0.7\textwidth]{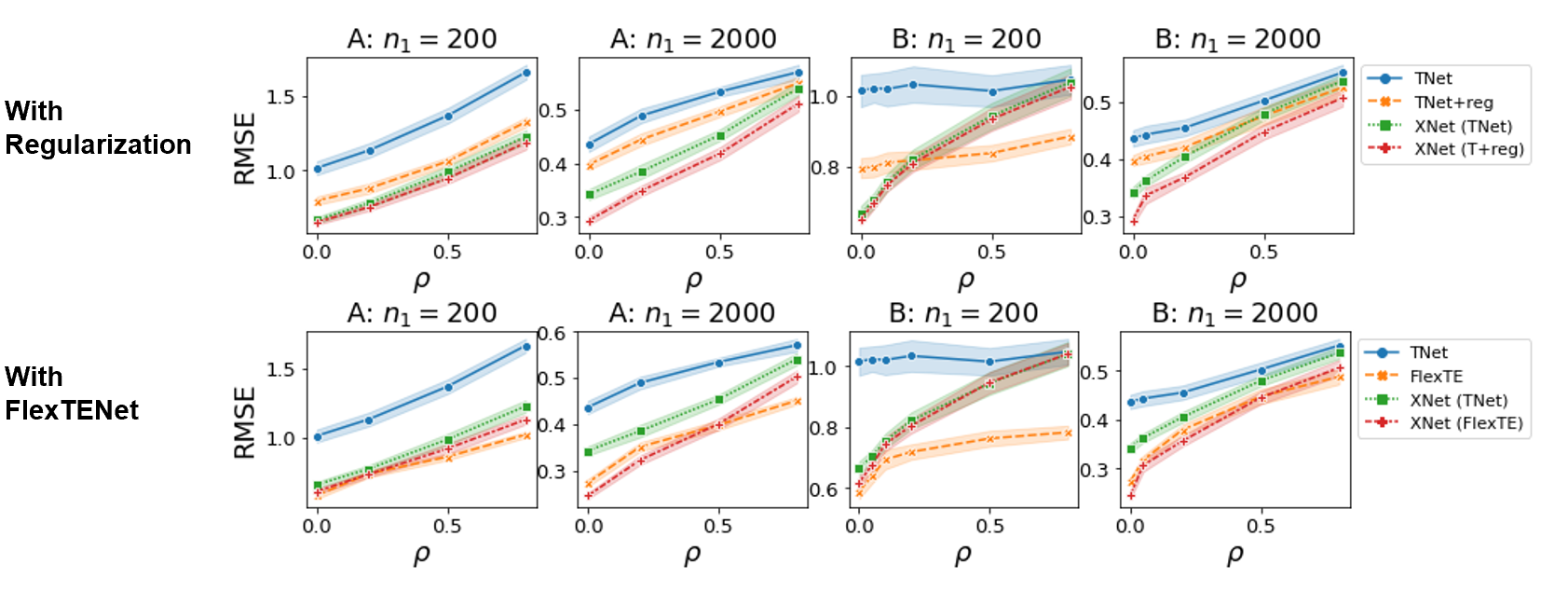}
    \vspace{-0.05in}
    \caption{RMSE of CATE estimation for the X-learner using different methods as first-stage estimators, by $\rho$ for different $n_1$ with $n_0=2000$ for setup A \& B. Avg. across 10 runs, one standard error shaded.}
    \label{fig:xres}
\end{figure}

\begin{figure}[!h]
    \centering
    \includegraphics[width=0.7\textwidth]{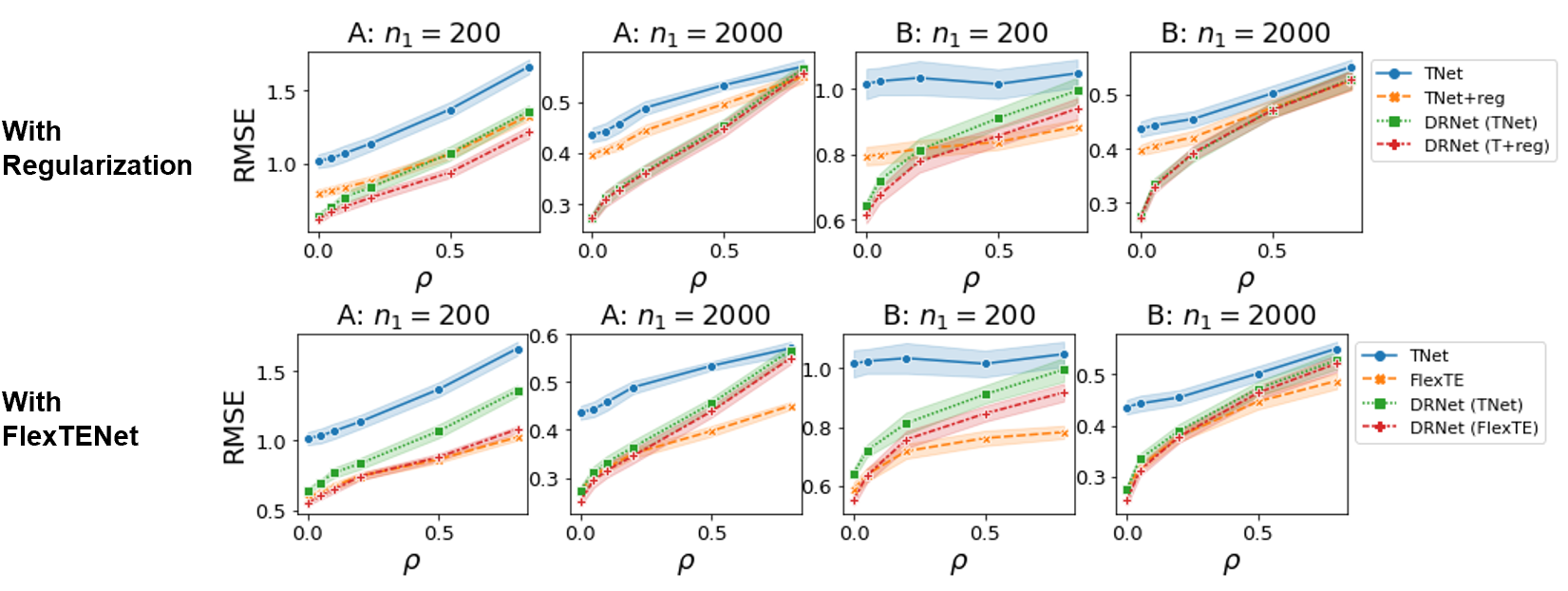}
    \vspace{-0.05in}
    \caption{RMSE of CATE estimation for the DR-learner using different methods as first-stage estimators, by $\rho$ for different $n_1$ with $n_0=2000$ for setup A \& B. Avg. across 10 runs, one standard error shaded.}
    \label{fig:drres}
\end{figure}

While meta-learners are usually implemented using vanilla first-stage estimators (i.e. in our case a separate neural network for each nuisance estimation task), we investigate here whether our improved indirect estimators can be used to improve performance of two-stage learners, here X- and DR-learner. In Fig. \ref{fig:xres} for the X-learner and Fig. \ref{fig:drres} for the DR-learner, we find that, while there are some performance increases, improvements in the second stage are much smaller than improvements to the first stage and, for the case of FlexTENet, the second stage does not improve upon the first-stage estimate. As discussed in section \ref{sec:implementation}, we did not use sample-splitting in our experiments as we found this to deteriorate performance for all meta-learners (especially in smaller sample sizes); this might be one reason for this finding.

\newpage
\subsection{Scaling of IHDP results (setups C \& D)}
\begin{figure}[!h]
    \centering
    \includegraphics[width=\textwidth]{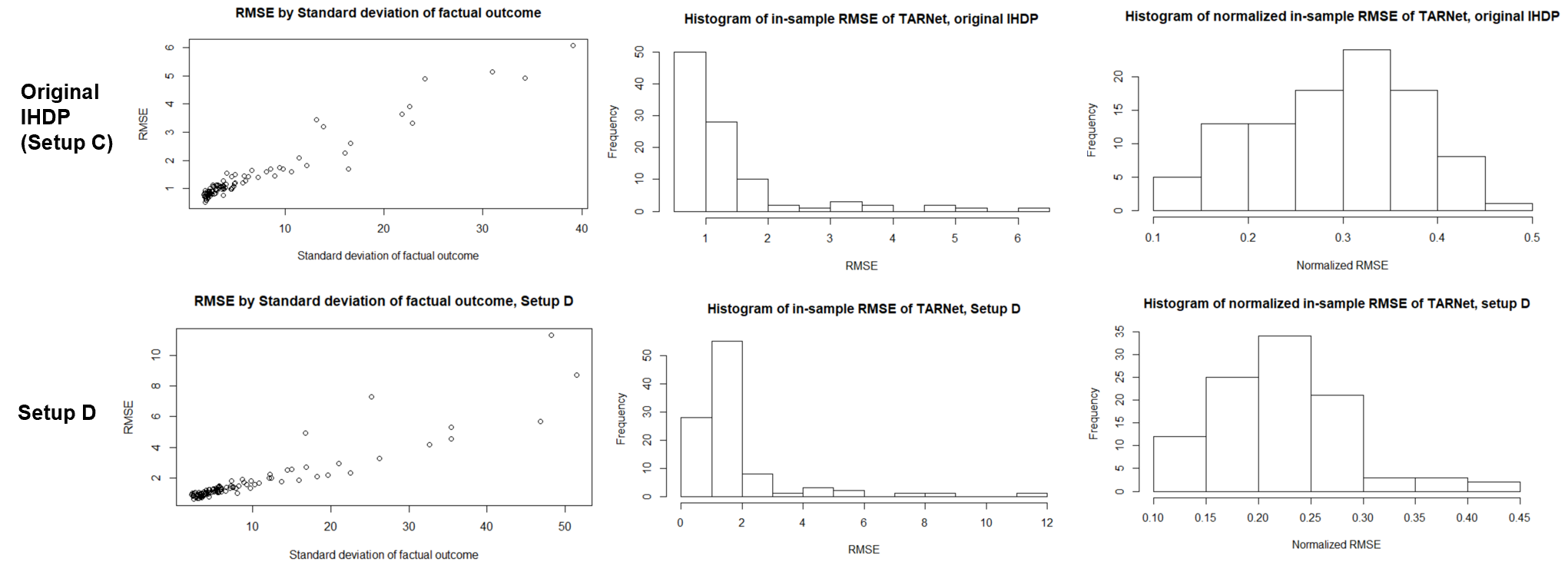}
    \vspace{-0.2in}
    \caption{In-sample RMSE of TARNet by standard deviation of factual outcomes in training sample, histogram of in-sample RMSE across runs and histogram for normalized in-sample RMSE across runs for Setup C (top) and Setup D (bottom)}
    \label{fig:ihdpvar}
\end{figure}

In setups C \& D, we observed that the scale of RMSE of CATE estimation varied by orders of magnitude across different runs of the simulation due to the exponential regression specification in the response surface, making performance in terms of RMSE incomparable across runs. We found that by averaging RMSE across runs, the relative performance was dominated by runs with high variance in factual outcomes (which arise in runs in which many variables enter the exponential specification). Therefore, we report RMSE normalized by standard deviation of the observed factual training data in the main text. As Figure \ref{fig:ihdpvar} highlights for the example of TARNet, this leads to much more well-behaved distributions of RMSE in both setups. 

In Table \ref{table:ihdp_unscaled} we report unnormalized results on the two IHDP setups for completeness. The results for TARNet differ from those reported in \cite{shalit2017estimating} for three main reasons: First, we used the IHDP-100 benchmark, and not the 1000 replications used in \cite{shalit2017estimating}. Second, as we highlighted above, the RMSE scores are clearly not normally distributed with the same mean such that some runs have much larger influence than others, and reported standard errors do not necessarily reflect the right confidence levels. Third, we used our own implementations and hyperparameter settings/architectural specification for these experiments, which differ slightly from those used in \cite{shalit2017estimating}.

\begin{table}[!h]
\centering
\vskip -0.1in
\caption{Unnormalized in- and out-of-sample RMSE of CATE estimation on the two IHDP setups. Averaged across 100 runs, standard error in parentheses.}\label{table:ihdp_unscaled}
\begin{tabular}{lllll}
\toprule
             & C, in         & C, out        & D, in         & D, out        \\ \midrule
TNet         & 1.572 (0.134) & 1.775 (0.198) & 2.857 (0.438) & 2.854 (0.445) \\
TNet + reg   & 1.418 (0.107) & 1.720 (0.193) & 2.355 (0.285) & 2.382 (0.293) \\
DR (+TNet)   & 1.681 (0.134) & 1.924 (0.198) & 1.764 (0.171) & 1.740 (0.158) \\
TARNet       & 1.384 (0.107) & 1.690 (0.196) & 1.772 (0.188) & 1.804 (0.204) \\
TARNet + reg & 1.350 (0.107) & 1.657 (0.197) & 1.623 (0.164) & 1.647 (0.181) \\
DR (+TARNet) &1.574 (0.125)& 1.801 (0.183)& 1.463 (0.128)& 1.448 (0.121)\\
Reparam.     & 2.136 (0.219) & 2.297 (0.269) & 1.394 (0.098) & 1.392 (0.092) \\
FlexTENet     & 1.226 (0.099) & 1.536 (0.182) & 1.966 (0.239) & 2.057 (0.261)\\
 \bottomrule
\end{tabular}
\end{table}
\newpage
\section{Additional benchmark datasets}
Below we present results on additional benchmark datasets: in section E.1 we consider the original response surfaces simulated for ACIC2016, and in section E.2 we consider performance on the Twins dataset with real outcomes. Throughout, we use the same hyperparameters as for setups A \& B.
\subsection{Original ACIC2016} As an additional benchmark dataset, we consider performance on the original simulations of ACIC2016 \cite{dorie2019automated}. The 77 settings vary in the complexity of response surfaces and the degrees of confounding, overlap and TE heterogeneity. They are based on the same covariates as simulations A and B, but differ in treatment assignment and response surfaces. In the competition, covariates were provided to participants without preprocessing, hence we start with the unprocessed dataset and, similar to \cite{assaad2020counterfactual}, we standardize all covariates and drop the three categorical variables as none of the considered methods are well-suited to handle them. We report results on 20\% test-sets for each of the 77 settings (averaging across 10 repetitions each). 

\begin{figure}[!h]
    \centering
    \includegraphics[width=\textwidth]{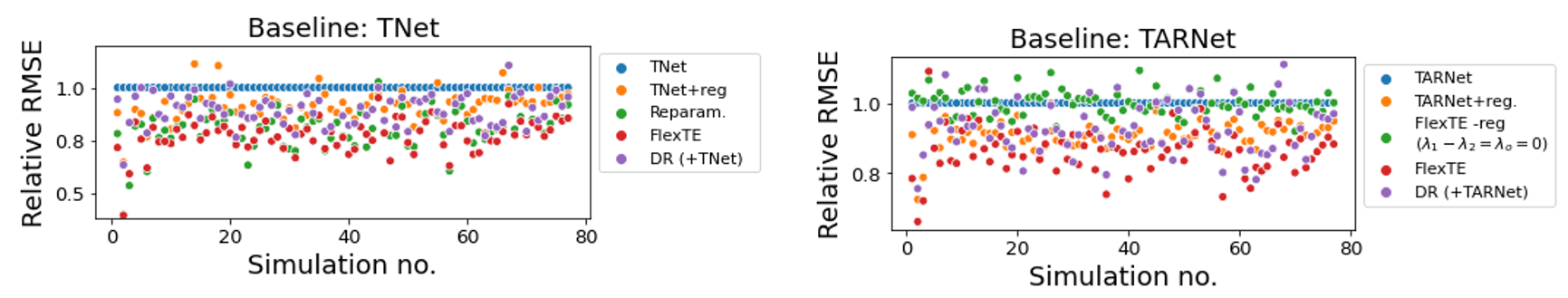}
    \vspace{-0.2in}
    \caption{RMSE relative to TNet (left) and TARNet (right) on the 77 simulation settings from ACIC2016. Averaged across 10 runs each.}
    \label{fig:acic}
\end{figure}

 In Fig. \ref{fig:acic}, we present results. Due to high variation in RMSE across the simulation runs (even within the same setting), we report RMSE(method)/RMSE(baseline) for TNet and TARNet as baseline. We note that our original findings on relative performance of the approaches from the main text hold up across a majority of settings. Apart from that,  we found little meaningful insights into sources of performance variation across different settings, possibly due to the very high variation in simulated response surfaces within and across settings. 
 
\subsection{Twins}
Evaluation of treatment effect estimators on real data is usually prohibited by the absence of ground-truth treatment effects and counterfactuals in practice. Twin studies in which each twin is assigned a different treatment therefore present an interesting exception: under the assumption of equivalence between two twins, realisations of both potential outcomes are observed. The Twins dataset considered by \cite{louizos2017causal, yoon2018ganite} contains one-year mortality outcomes for 11400 pairs of twins with 39 relevant covariates\footnote{We obtained the preprocessed dataset used in \cite{yoon2018ganite} from the authors, which is derived from the data provided by \cite{almond2005costs}}. Here, the treatment is `being heavier at birth', so that this data can be used to evaluate the effect of birthweight on infant mortality.

The outcome in this dataset is binary and (fortunately) imbalanced; mortality rates over the full data are $16.1\%$ and $17.7\%$ for treated and untreated, respectively. Due to the binary nature of the data and this imbalance, the signal for the presence of treatment effects (which necessitates observing opposite outcomes for pairs of twins), is relatively weak and noisy. Therefore, we use a large test set and hold out 50\% (5700 pairs of twins) for testing. For training, we randomly select one twin from each pair with (constant) probability $p_{treat} \in \{0.1, 0.25, 0.5, 0.75, 0.9\}$ to investigate the performance of each method on imbalanced data as in the main text, but now with \textit{real} data. Additionally, we vary the number of training examples $n_{train}\in \{500, 1000, 2000, 4000, 5700\}$ to assess the sample efficiency of different methods; leading to 25 different settings considered for the Twins data.

\textbf{Metrics} As the true $\tau(x)$ and $\mu_w(x)$ are unobserved, we can only use realisations of $Y(w)$ to evaluate all models. First, as \cite{yoon2018ganite} we consider the RMSE on the observed counterfactual difference, i.e. $\sqrt{\frac{1}{n} \sum^n_{i=1}\big((y_i(1)-y_i(0))- (\hat{\mu}_1(x_i)-\hat{\mu}_0) \big)^2}$, as a metric to evaluate the quality of the treatment effect estimate. Because $Y$ is binary, $y_i(1)-y_i(0) \in \{-1, 0, 1\}$ while $\hat{\mu}_1(x_i)-\hat{\mu}_0$ is not, and this metric is very noisy. Therefore, we additionally consider $Y(1)-Y(0)$ as the target in a 3-class classification problem, where
\begin{equation}\label{iteclass}
    \mathbb{P}(Y(1)-Y(0)=t|X=x)=\begin{cases} \mu_0(x) \times (1-\mu_1(x)) & \text{ if }~ t=-1\\
     (1-\mu_0(x)) \times \mu_1(x) &\text{ if }~t=1\\
    \mu_0(x)\times \mu_1(x) + (1-\mu_0(x) \times (1-\mu_1(x)) &\text{ if } ~t=0
    \end{cases}
\end{equation}
if we assume that the two potential outcomes are conditionally independent. We can compute $\hat{\mathbb{P}}(Y(1)-Y(0)=t|X=x)$ for all models which predict potential outcomes using (\ref{iteclass}), and evaluate its fit on the real data using standard classification metrics. Here, we report the area under the receiver-operating curve (henceforth: AUC). Further,  similar to \cite{louizos2017causal}, we also evaluate predictive performance on each of the POs separately through the AUC.

\subsubsection{Performance on estimating CATE}
\begin{figure}[!h]
    \centering
    \includegraphics[width=\textwidth]{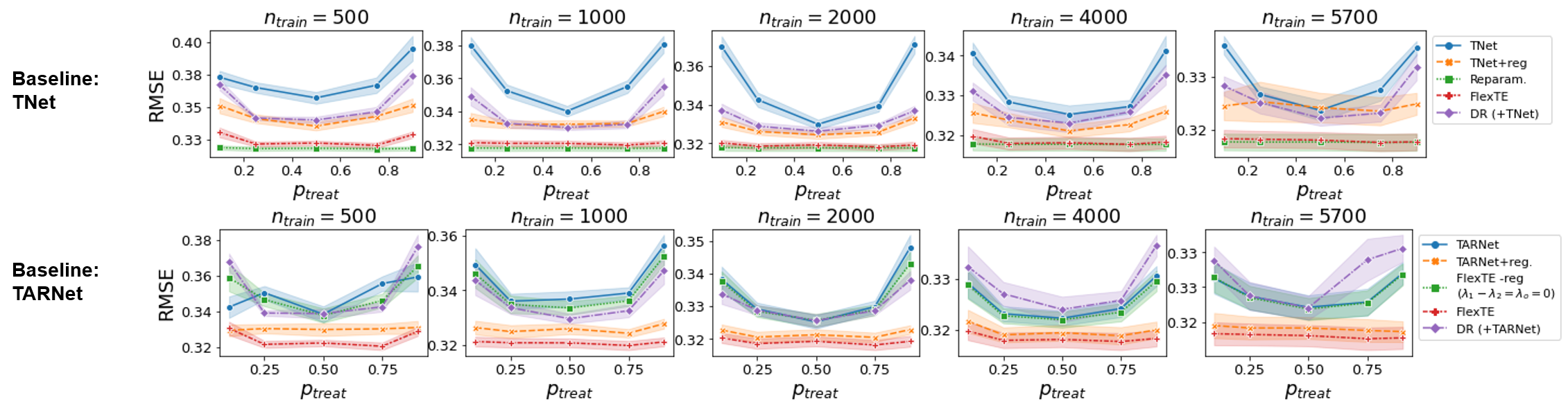}
    \vspace{-0.05in}
    \caption{RMSE on the counterfactual difference (lower is better), by $p_{treat}$ for different $n_{train}$ using TNet (top) and TARNet (bottom) as baseline. Avg. across 10 runs, one standard error shaded.}
    \label{fig:twins_rmse}
\end{figure}

\begin{figure}[!h]
    \centering
    \includegraphics[width=\textwidth]{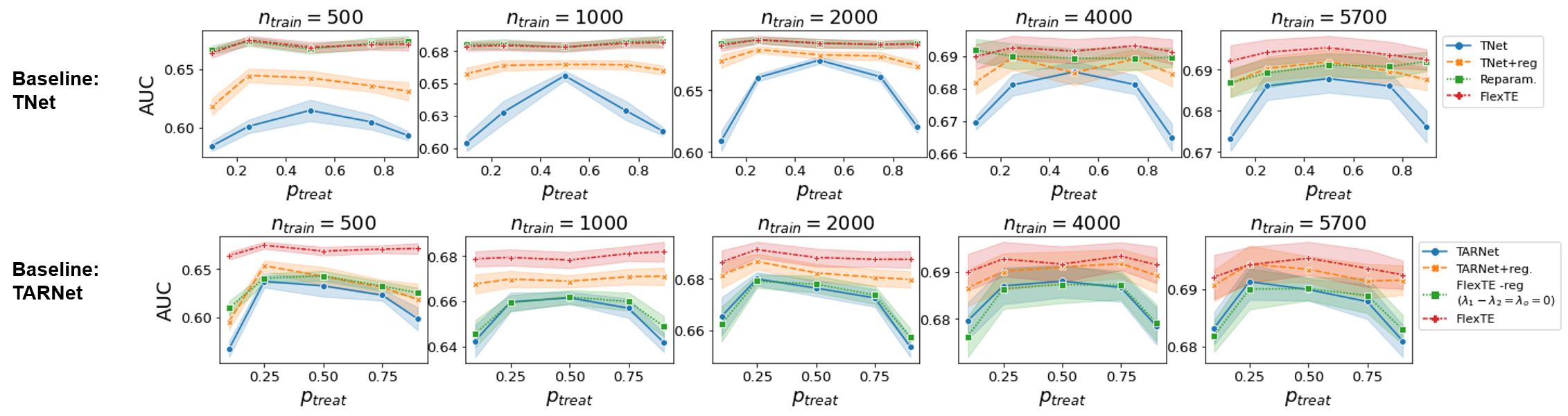}
    \vspace{-0.05in}
    \caption{AUC on the counterfactual difference (higher is better), by $p_{treat}$ for different $n_{train}$ using TNet (top) and TARNet (bottom) as baseline. Avg. across 10 runs, one standard error shaded.}
    \label{fig:twins_auc_ite}
\end{figure}

In Fig. \ref{fig:twins_rmse} and Fig. \ref{fig:twins_auc_ite}, we present results on performance of estimating the counterfactual difference, measured by RMSE and AUC, respectively. We observe that Reparametrization and FlexTENet perform almost equivalently and best throughout as measured by both metrics. Both are most robust to changes in $p_{treat}$, and, contrary to most other methods, perform close to optimal already with only 500 samples. Further, the regularization approach improves upon its baselines also in this setting. The conclusions on relative performance made in the main text thus largely hold up also here; the main difference is that here FlexTENet performs very well also for small $n_{train}$. This, and the near-equivalence of FlexTENet and reparametrization approach,  might indicate that there is both significant shared structure \textit{and} relevant (\textit{additive}) heterogeneity between the (unknown) POs in the Twins dataset. 

\subsubsection{Performance on estimating the POs} 
\begin{figure}[!h]
    \centering
    \includegraphics[width=\textwidth]{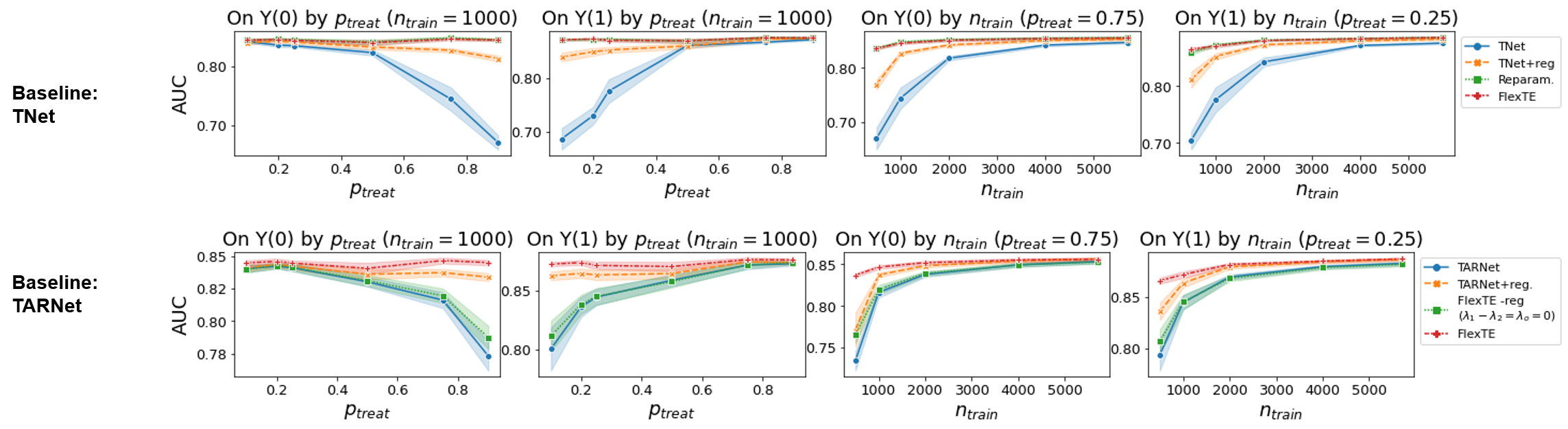}
    \vspace{-0.05in}
    \caption{AUC on the potential outcomes (higher is better), by $p_{treat}$ (left) and by  $n_{train}$ (right) using TNet (top) and TARNet (bottom) as baseline. Avg. across 10 runs, one standard error shaded.}
    \label{fig:twins_auc_po}
\end{figure}

Finally, we consider performance on estimating the POs separately in Fig. \ref{fig:twins_auc_po}. In the left panels we plot AUC on each potential outcome for different levels of $p_{treat}$, and observe that all approaches can significantly improve the performance on estimating the POs when there is imbalanced treatment assignment; most likely this is because they provide additional supervision for the underrepresented treatment arm. In the right panels we plot AUC for underrepresented treatment arms by different levels of $n_{train}$, and observe that FlexTENet and the reparametrization approach provide such supervision most sample-efficiently: they reach near-optimal performance with a fraction of the available samples.

\section{NeurIPS Checklist}

\begin{enumerate}

\item For all authors...
\begin{enumerate}
  \item Do the main claims made in the abstract and introduction accurately reflect the paper's contributions and scope?
    \answerYes{}
  \item Did you describe the limitations of your work?
    \answerYes{(1) In Section 2.1, we discuss that the ability to interpret treatment effects as causal always relies on assessment of the plausibility of assumptions, which should be conducted by a domain expert. (2) We highlight that the conclusions we make here using neural networks should be consolidated using other base-methods in future work.}
  \item Did you discuss any potential negative societal impacts of your work?
    \answerYes{See answer (1) above. The limitation of all methods trying to infer causal effects from observational data is the presence of strong identifying assumptions; if such assumptions are not properly assessed by domain experts prior to deployment, the conclusions drawn using such methods may be misleading. }
  \item Have you read the ethics review guidelines and ensured that your paper conforms to them?
    \answerYes{}
\end{enumerate}

\item If you are including theoretical results...
\begin{enumerate}
  \item Did you state the full set of assumptions of all theoretical results?
    \answerNA{}
	\item Did you include complete proofs of all theoretical results?
    \answerNA{}
\end{enumerate}

\item If you ran experiments...
\begin{enumerate}
  \item Did you include the code, data, and instructions needed to reproduce the main experimental results (either in the supplemental material or as a URL)?
    \answerYes{Code is provided at \url{https://github.com/AliciaCurth/CATENets}}
  \item Did you specify all the training details (e.g., data splits, hyperparameters, how they were chosen)?
    \answerYes{All training details are specified in appendix C, and provided within the code at  \url{https://github.com/AliciaCurth/CATENets}}
	\item Did you report error bars (e.g., with respect to the random seed after running experiments multiple times)?
   \answerYes{}
	\item Did you include the total amount of compute and the type of resources used (e.g., type of GPUs, internal cluster, or cloud provider)?
   \answerYes{All training details are specified in appendix C}
\end{enumerate}

\item If you are using existing assets (e.g., code, data, models) or curating/releasing new assets...
\begin{enumerate}
  \item If your work uses existing assets, did you cite the creators?
    \answerYes{All sources of data are listed in appendix C}
  \item Did you mention the license of the assets?
    \answerNA{No explicit license was provided with the data.}
  \item Did you include any new assets either in the supplemental material or as a URL?
    \answerYes{All code is provided in the supplement.}
  \item Did you discuss whether and how consent was obtained from people whose data you're using/curating?
    \answerNA{Data is publicly available.}
  \item Did you discuss whether the data you are using/curating contains personally identifiable information or offensive content?
    \answerNA{Data was previously de-identified}
\end{enumerate}

\item If you used crowdsourcing or conducted research with human subjects...
\begin{enumerate}
  \item Did you include the full text of instructions given to participants and screenshots, if applicable?
    \answerNA{}
  \item Did you describe any potential participant risks, with links to Institutional Review Board (IRB) approvals, if applicable?
    \answerNA{}
  \item Did you include the estimated hourly wage paid to participants and the total amount spent on participant compensation?
    \answerNA{}
\end{enumerate}

\end{enumerate}


\end{document}